\documentclass[runningheads]{llncs}
\usepackage{graphicx}
\usepackage{comment}
\usepackage{amsmath,amssymb} %
\usepackage{color}

\usepackage[width=122mm,left=12mm,paperwidth=146mm,height=193mm,top=12mm,paperheight=217mm]{geometry}

\usepackage{epsfig}
\usepackage{microtype}

\usepackage{booktabs}

\usepackage{caption} %

\newcommand{\etal}{\textit{et al}.}

\newbox\jsavebox

\usepackage[pagebackref=true,breaklinks=true,letterpaper=true,colorlinks,bookmarks=false]{hyperref}

\begin{document}
\pagestyle{headings}
\mainmatter
\def\ECCVSubNumber{2946}  %

\title{Hidden Footprints: Learning Contextual Walkability from 3D Human Trails}

\titlerunning{Hidden Footprints: Learning Contextual Walkability from 3D Human Trails}
\author{Jin Sun \and Hadar Averbuch-Elor \and Qianqian Wang \and Noah Snavely}
\authorrunning{J. Sun, H. Averbuch-Elor, Q. Wang, and N. Snavely}
\institute{Cornell Tech, New York, NY 10044\\
\email{\{jinsun,hadarelor,qw246,snavely\}@cornell.edu}}
\maketitle

\begin{figure}
\includegraphics[width=\textwidth]{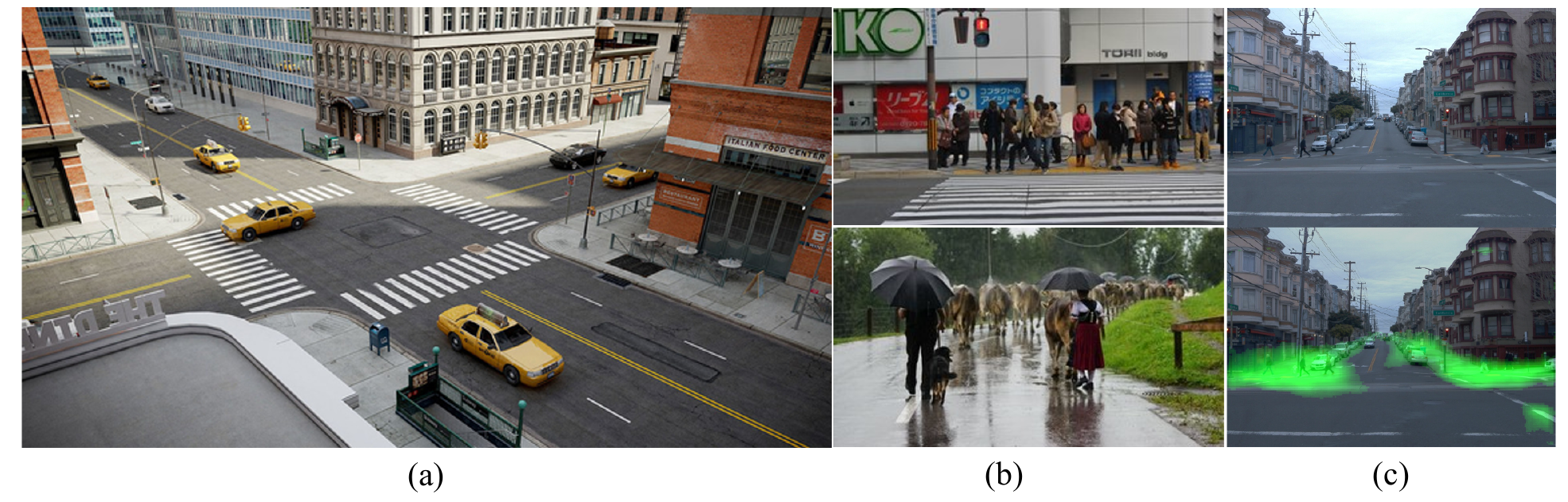}
\caption{(a) Imagine an urban scene like this one. Where could people walk in such a scene? 
The answer is not simply `crosswalk' or `sidewalk'---scene context plays a significant role.
For example, people can be standing on sidewalks and not on crosswalks while waiting for a green light (b, top). 
A road can be open to walk, or only partially walkable (e.g., when occupied by cows, as in (b, bottom)). 
(c) We propose to predict where people could walk (highlighted in green) by learning from observing human behavior in weakly annotated data.
 }
\label{fig:teaser}
\end{figure}

\begin{abstract}
Predicting where people can walk in a scene is important for many tasks, including autonomous driving systems and human behavior analysis. Yet learning a computational model for this purpose is challenging due to semantic ambiguity and a lack of labeled data: current datasets only tell you where people \textit{are}, not where they \textit{could be}. We tackle this problem by leveraging information from existing datasets, without additional labeling. We first augment the set of valid, labeled walkable regions by propagating person observations between images, utilizing 3D information 
to create what we call \textit{hidden footprints}. However, this augmented data is still sparse. We devise a training strategy designed for such sparse labels, combining a class-balanced classification loss with a contextual adversarial loss. Using this strategy, we demonstrate a model that learns to predict a walkability map from a single image.
We evaluate our model on the Waymo and Cityscapes datasets, demonstrating superior performance compared to baselines and state-of-the-art models.
\end{abstract}

\section{Introduction}
Walking, one of our most 
common daily activities, requires complex scene understanding and planning.
Consider the image in Figure~\ref{fig:teaser}(a). We have no trouble imagining where a person might walk in this scene---on sidewalks, in crosswalks, or into a subway entrance.
Given an image, predicting where people might walk is critical for engineering robust autonomous systems like self-driving cars that must anticipate human behaviors.
Such a capability could also be used in urban planning and public policy making to assess walkability in a neighborhood~\cite{walkability}. 

At first glance, 
one might think that 
predicting where people might appear in an image is
equivalent to reasoning about semantics, since pedestrians are associated with sidewalks but not streets, etc. Perhaps we can simply convert a semantic segmentation into a prediction of walkable regions. 
However, we find empirically that semantic categories are a surprisingly poor predictor of where humans might walk in a given scene.
Common semantic categories like road and sidewalk alone do not fully explain affordances that are indicated by other visual cues, such as crosswalks, curb ramps, and red lights.
And some cases require even more complex reasoning: 1) one may walk in front of a parked car, but not a moving one;
2) a sidewalk can become unwalkable due to temporary construction work;
3) during special events, a street that is usually unsafe for walking might be closed to traffic and hence walkable.
A single semantic label is insufficient for reasoning about these cases, and labeling scenes with such fine-grained semantic granularity as `street during a special event' would be prohibitively expensive.

In this work, we propose to predict \textit{contextual walkability}---a map of where human \textit{could} be walking on under the current scene's conditions---to distinguish from measuring a generic static `walkable' map defined by semantics or geometry.
We seek to design a computational model to predict such a map given a street image.
Rather than learning from a pre-defined set of semantic categories and detecting where objects \textit{are}, we address this novel task of predicting where they \emph{could be} by taking a data-driven 
approach where we learn from direct observation of human behavior.
While this approach allows for a flexible modeling of context, this task is challenging because the visual cues for predicting where people \emph{could} be are significantly more ambiguous. 
Furthermore, 
no such labeled dataset exists to the best of our knowledge.

\begin{figure*}[tb]
  \centering
    \includegraphics[width=\linewidth]{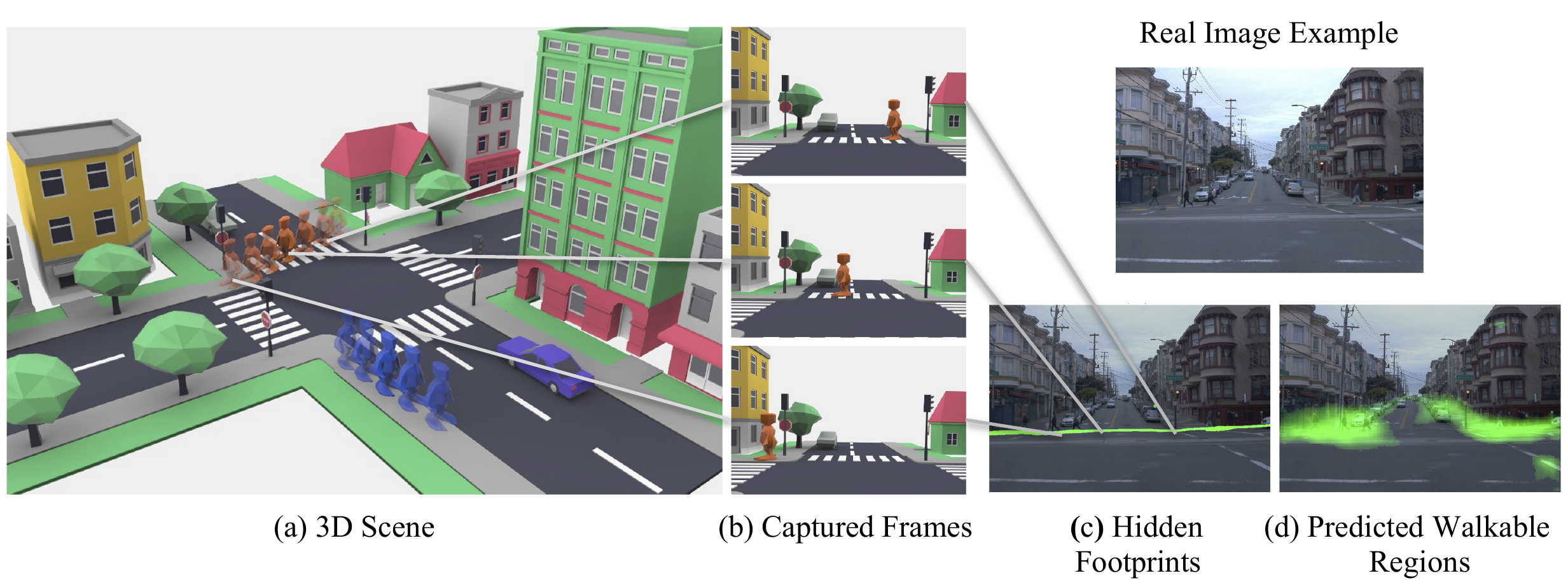}
  \captionof{figure}{When a person is observed walking through a scene (a), a sequence of images (b) of that person defines a 
  3D trail, where each position indicates a valid location for people to walk in the scene. We call such virtual trails \emph{hidden footprints} (c). 
While such trails provide partial annotation of walkable regions, our trained model can generalize to other plausible locations (d).
  }
  \label{fig:intro}
\end{figure*}

Fortunately, 
we are witnessing a surge of street scene datasets collected for autonomous driving, such as the Waymo~\cite{waymo_open_dataset}, nuScenes~\cite{nuscenes2019}, and Argoverse~\cite{chang2019argoverse} datasets.
In such datasets, a car drives around urban areas and continuously captures sensory data, including images.
These datasets also 
provide 2D and 3D people labels,
along with the 3D poses of the vehicle and its cameras.
We leverage such data for learning contextual walkability models 
\emph{without additional labels}.
Our approach, illustrated in Figure~\ref{fig:intro}, relies on two key insights:
\begin{itemize}
\item We can significantly augment walkable region labels in a scene by re-projecting all 3D people annotations into a given reference frame using the provided camera poses (Section~\ref{sec:projection}). 
Such ``hidden footprints'', which capture true walkable regions, not only expand the labeling of walkable regions, 
but also facilitate the learning of purely contextual cues because no real person is present at those projected locations. 
\item However, the number of people we observe in a scene is still limited: many walkable regions that share similar contextual cues may be missing from the ground-truth. 
We thus train a dense prediction neural network using a novel training strategy to recover a walkability distribution from under-sampled ground-truth labels (Section~\ref{sec:distribution}).
Our training objective guides a model to not only be faithful to the original ground-truth labels, but also to be able to expand beyond to contextually similar regions in the same scene.
\end{itemize}
Combining these two insights yields a model that can predict a broad distribution with good coverage of where people could walk, trained solely from an initially sparse and incomplete set of true labels.

We evaluate our approach on popular datasets that feature pedestrians (Section~\ref{sec:experiment}).
Our model significantly outperforms alternative approaches, in terms of covering true walkable locations and classification performance on a manually labeled test set.
In addition, our model generalizes surprisingly well to the Cityscapes dataset, even with no fine-tuning.
Finally, we show that our hidden footprints framework is flexible and can be adapted to predict additional maps, such as expected pedestrian walking directions.

\section{Related work}

Our work addresses
contextually-correct placement of 
objects
within a target image. Sun and Jacobs \cite{sun2017seeing} learn a representation of context for locating missing objects in an image, such as curb ramps on sidewalks that ought to be present but are not. 
Like us, Chien \etal~\cite{chien2017detecting} detect non-existent pedestrians. However, unlike our 3D-aware data collection approach, they use images with labeled pedestrians and remove a subset of them using image inpainting. Lee \etal~\cite{lee2018context} use semantic maps to jointly predict the locations and shapes of potential people. Our work doesn't require semantic inputs.

Several prior works 
focus on synthesizing pedestrians in images. Pedestrian-Synthesis-GAN~\cite{ouyang2018pedestrian} trains a GAN on pedestrian bounding boxes filled with random noise. The network learns to synthesize the missing pedestrian based on nearby regions. Lee \etal~\cite{lee2019inserting} address the related problem of cutting an object from one video, e.g., a walking pedestrian, and pasting it into a user-specified region in another video. Hong \etal~\cite{hong2018learning} use semantic maps to infer a person's shape and appearance inside user-provided bounding boxes.

Inserting an object into a photo is useful for data augmentation, and determining where to insert an object is often an essential step.
ST-GAN~\cite{lin2018st} uses a GAN framework equipped with a spatial transformer network to output a natural composite restricted to the space of target geometric manipulations.
Dwibedi \etal~\cite{dwibedi2017cut} propose a simple approach to synthesizing training images for instance detection tasks.
Huang \etal~\cite{huang2017expecting} focus on 
detecting pedestrians in unexpected scenarios and use GANs to select a realistic subset of synthetically generated images.
Interactive techniques for object insertion are also common. Lalonde \etal~\cite{lalonde2007photo} built an interactive system that, given a %
target image and a desired 3D location in the scene, retrieves object instances for insertion. 
Our model can aid such methods by providing good candidate object locations.

Our work is also related to human-centric scene analysis.
Kitani \etal~\cite{kitani2012activity} combine semantic scene understanding and optimal control theory to forecast the activities of people in an image.
Fouhey \etal~\cite{Fouhey12} use human actions as a cue for 3D scene understanding.
A key subproblem of human analysis is to infer affordances, i.e., potential interactions of people with environments.
Gupta \etal~\cite{gupta20113d} manually associate human actions with exemplar poses and use 3D correlation between poses and scenes to obtain valid locations for these actions in the scene. 
Wang \etal~\cite{wang2017binge} use human poses extracted from TV sitcoms to learn affordances.
Li \etal~\cite{li2019putting} further encourage physically feasible solutions by leveraging geometric knowledge from voxel representations.
Chuang \etal~\cite{chuang2018learning} train a model with semantic maps to reason about %
affordances that respect both physical rules and social norms.
During inference, our model takes a single image as input and requires no additional information such as 3D or semantics.

\begin{figure*}[tb]
  \centering
  \includegraphics[width=0.9\textwidth]{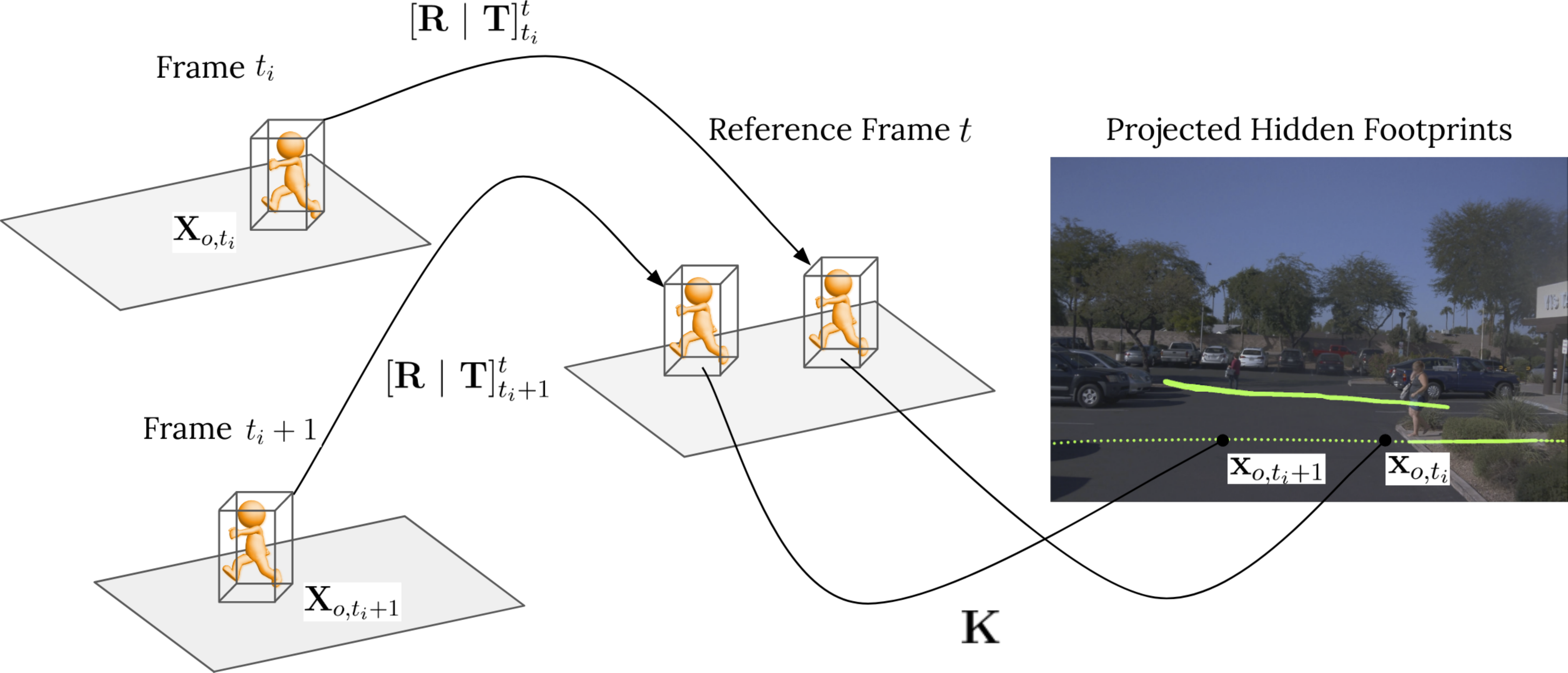}
  \caption{\textit{Propagating 3D annotations to new frames.} Each labeled person $o$ is represented by the 3D centroid, $\mathbf{X}_{o,t_i} \in \mathbb{R}^{3}$, of the bottom face of its 3D cuboid (representing the averaged feet location) in the coordinate frame of frame $t_i$. This point is transformed to the reference frame $t$'s coordinate systems by the 3D rotation and translation $[\mathbf{R}~|~\mathbf{T}]_{t_i}^t$ relating these two frames. The transformed centroid is then projected to the image plane via the camera intrinsics matrix $\mathbf{K}$. The process continues for $i+1, i+2, \ldots, T$ until the end of the  sequence. The final reference image plane contains the accumulation of these virtual 2D footprints from the whole sequence.}
    \label{fig:proj}
\end{figure*}

\section{Propagating hidden footprints}\label{sec:projection}
In this section, we describe 
how to obtain
an augmented set of valid walkable locations in a scene,
exploiting the nature of data collection 
in recent
datasets.
We focus on the Waymo Open Dataset \cite{waymo_open_dataset}, but any dataset that provides 3D labels and camera parameters, such as the nuScenes \cite{nuscenes2019}, could be used as well.

In autonomous driving datasets, street scenes are often captured in units of driving sequences, that is, sequences of frames each with a known 3D pose in a world coordinate system.
The 3D location of each visible object (e.g., cars, persons, signs, and cyclists) is annotated.
Using camera poses (and associated intrinsics), any 3D location can be projected to any frame in the same sequence.

We observe that when a person is walking during the capture,
the ``footprints'' they leave are not only valid locations for a person to walk on for the \emph{current} frame, but also a valid location for \emph{any other frame} in the same scene.

Therefore, we project all person locations, across the entire sequence, from 3D coordinates to 2D coordinates in all frames.
This results in a massive expansion of walkable location labels, and, most importantly, these locations are all real, not hallucinated. 
Note that we do not project labels across sequences, as walkability might change over longer periods of time (e.g., from day to night).
We further assume that walkability conditions remain the same during the duration of the relative short sequences (around 20 seconds in Waymo).

Figure \ref{fig:proj} illustrates our process for propagating these ``hidden footprints''.
In particular, we generate a footprint map $L$ for each frame $t$ from all time steps and all objects in the same driving sequence:
\begin{equation}\label{eq:proj}
L_t(\mathbf{x}) = \sum_{i=1}^T \sum_{o=1}^O  g(\mathbf{x}_{o,t_i} - \mathbf{x}; \sigma),~~~~~\text{where}~~ \mathbf{x}_{o,t_i}=\mathbf{\Pi}_{t_i}^{t} (\mathbf{X}_{o,t_i}),
\end{equation}
$T$ is the total number of frames in the sequence, 
$O$ is the number of objects (e.g., people) present in the scene, 
$\mathbf{x}$ is a 2D location in $L_t$ in the reference frame $t$, 
$g(\cdot; \sigma)$ is a Gaussian kernel function with standard deviation $\sigma$, 
and $\mathbf{\Pi}_{t_i}^t$ is a transformation from frame $t_i$ to image coordinates in frame $t$, consisting of a camera rotation $\mathbf{R}$ and translation $\mathbf{T}$, followed by an application of the camera intrinsics matrix $\mathbf{K}$ such that 
$\mathbf{\Pi}_{t_i}^t = \mathbf{K}[\mathbf{R}~|~\mathbf{T}]_{t_i}^t$. 
In other words, $\mathbf{\Pi}_{t_i}^{t}$ 
maps a 3D point $\mathbf{X}_{o,t_i}$ to a 2D point $\mathbf{x}_{o,t_i}$. 
(For simplicity, we do not distinguish between Cartesian and homogeneous coordinates.)
The accumulated map shows the observed contextual walkability in the reference scene.
The benefits of our hidden footprints are two-fold:

\smallskip
\noindent\textbf{Vastly more labels.}
The propagated locations represent a significant expansion of the labels from existing pedestrians.
All these locations are real and valid, not artificially augmented.
Such abundant labeled data greatly improve a learned model's generalization power.

\smallskip
\noindent\textbf{Easier to learn human context.}
Since the propagated locations usually do not contain an existing person in the reference frame,
no special treatment is necessary in order to learn a context-only model for human walkability, and hence a network trained on these augmented labels will not simply learn to detect people.
In prior work, an object of interest has to be either inpainted with background \cite{chien2017detecting} or blocked \cite{sun2017seeing} to learn a context-only model.

\smallskip
\noindent By propagating hidden footprints,
we obtain a large expansion in the available training labels for learning contextual walkable locations.
However, it still does not cover \emph{all} possible walkable locations.
For example, there might be people walking on a sidewalk on one side of a street in a given sequence, while the other side is entirely empty. Both are likely to be equally valid walkable locations given their contextual similarity.
A good model should 
predict both sides of the sidewalk as contextually walkable.
In the next section we describe a training strategy to obtain such a 
model.

\section{Contextual walkability from partial observations}\label{sec:distribution}

\begin{figure*}[tb]
  \centering
  \includegraphics[height=1.8in]{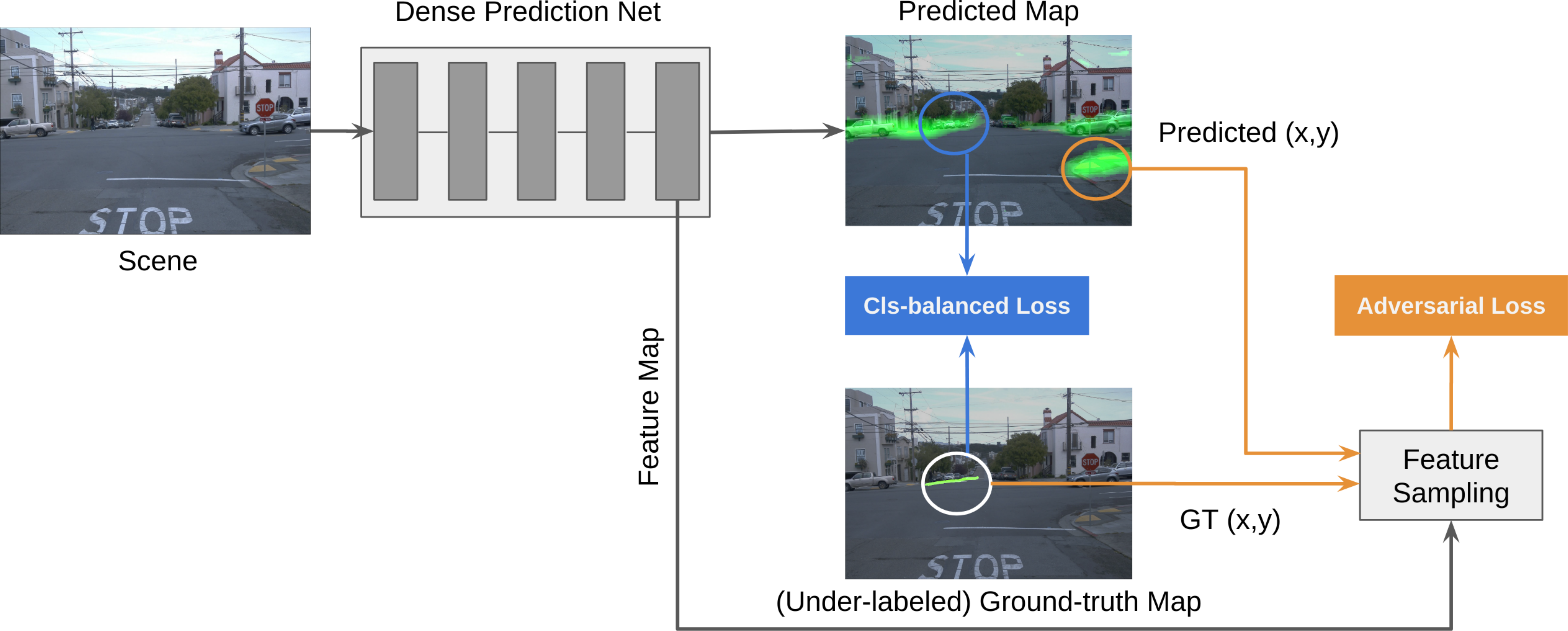}
  \caption{We model the underlying ground-truth distribution by training a dense prediction network with two losses. The first is a \emph{class-balanced classification loss} that encourages the network to predict positives (blue circle) for ground-truth positive regions (white circle). The second is an \emph{adversarial feature loss} used for determining hard false positive regions (orange circle). These regions will not be penalized if they share similar visual context with the true positive regions. }
    \label{fig:dist}
\end{figure*}

Propagating hidden footprints greatly increases the number of observations, alleviating, but not completely solving, our prediction task.
Our goal is to learn the latent underlying distribution of where a person can walk in a scene from partial observations.
We propose a 
training strategy with two losses that each focuses on different regions where ground-truth labels and predictions disagree.
Our approach is illustrated in Figure~\ref{fig:dist}.

In particular, we treat positive labels and negative labels differently.
For positive ground-truth locations (Section~\ref{sec:4-rce}), we discourage the model from predicting false negatives by using a class-balanced 
classification loss.
For negative ground-truth locations (Section~\ref{sec:4-fl}), we are more forgiving of discrepancies due to the fact that ground-truth is severely under-sampled (and hence many negatives in the ground-truth could in reality be walkable). 
We use deep visual features extracted at those ambiguous locations as a proxy to measure whether they are from the same distribution as ground-truth locations---i.e., contextually similar.
Together, these terms encourage the model to make faithful predictions for ground-truth locations and educated guesses in uncertain regions.

\subsection{Class-balanced classification loss for a global prediction}
\label{sec:4-rce}
We treat the prediction of contextual walkability of an image as a binary classification problem: given a location in an image, is it possible for a person to be walking there?
This setting is used in previous context modeling work \cite{sun2017seeing,chien2017detecting}.

Directly using a standard classification loss such as the cross-entropy loss is problematic, as it will be dominated by the overwhelmingly large and uncertain negative labels;
Waymo images have many times more background pixels than human ones.
We reduce such bias by adding class weights to the ground-truth labels.
In particular, for each image, we re-balance the positive and negative samples by the reciprocal of their label frequencies:
\begin{equation}
\mathit{CBL}(p, y) = - c_{\text{pos}} \cdot y \cdot \log(p) - c_{\text{neg}} \cdot (1-y) \cdot \log(1-p),
\end{equation}
where $p$ is the model's prediction, $y$ is a binary 1 or 0 label, and $c_{\text{pos}}$ and $c_{\text{neg}}$ are class weights for positive and negative labels, respectively.
We convert the footprint map $L$ from Eq. \ref{eq:proj} into a binary map by assigning non-zero values to 1.
By adding a higher weight to positive label locations, we encourage the model to make correct predictions on them.
Note that our overall training strategy is not limited to the specific choice of the class-rebalanced loss. 
We tried alternatives such as the Focal Loss~\cite{lin2017focal}, 
but found that they gave similar performance.

\subsection{Adversarial feature loss for hard false positives}
\label{sec:4-fl}
For this loss, we only investigate possible false positives and use an adversarial approach~\cite{goodfellow2014generative} to handle regions predicted as walkable by our model but where the ground-truth says otherwise.
We use features extracted from the same prediction model as a proxy to measure contextual similarity and encourage the model to produce positives at locations that 
are visually similar to known positives.%

More concretely, after the model outputs a prediction map, we sample locations where the predicted score is high but the ground-truth label is negative.
Ground-truth positives are sampled as well.
We then extract features at these locations, and feed them into a discriminator network.
While the classification loss penalizes false positives equally, the adversarial loss is lower for false positives that share similar features with ground-truth locations.
By training the discriminator to distinguish whether features are from the same distribution, gradients flow back to the walkability prediction network to encourage it to predict high scores more confidently on plausible locations.

To sample features,
we explored a few distribution sampling methods, including max sampling and rejection sampling.
To balance efficiency and effectiveness, we adopt a sampling strategy that selects the top $K\%$ scored regions and uniformly samples $N$ locations among them.
Though the sampling operation is not differentiable, gradients from the discriminator flow back to the model through the sampled features, and so the whole pipeline can be trained end-to-end.

\medskip
\noindent The class-balanced classification loss and the adversarial feature loss work together to encourage the network to faithfully predict 
ground-truth positive labels, 
while also expanding to locations that share similar visual context.

\medskip
\noindent\textbf{Discussion.} Compared to our proposed training strategy, there are alternatives for this task. Mean Squared Error (MSE) is a popular loss for modeling unimodal probability maps (e.g., \cite{newell2016stacked}).
We find that models trained with MSE produce visually noisier output.
This may be due to the multi-modal nature of walkability maps.
The training can also be done by an adversarial-only loss  \cite{chien2017detecting}, without any additional direct regression. 
We find that training is unstable without regression.
Our training can also be seen as a soft handling of unlabeled samples, compared to pseudo-labeling used in semi-supervised settings \cite{chapelle2009semi}. 
We empirically compare the proposed training strategy with the alternatives in the next section.
\section{Evaluation}\label{sec:experiment}

In this section, we evaluate how well our proposed approach
works as a contextual walkability model.
Just as our problem lacks full ground-truth for training, a proper evaluation is also non-trivial.
To help provide a comprehensive view of performance, our evaluation considers several aspects: 1) quality of predictions over the whole test set measured by the region \emph{expansion multiplier} over the labeled data; 2) a standard classification metric on a subset with manual ground-truth labels; 3) quality of predictions compared to semantics based reasoning and alternatives, and 4) generalization power to another dataset.
The results suggest that our method learns a good universal walkability model.
It greatly expands positive predictions from limited labels, makes reasonable additional predictions based on contextual similarity, and is able to generalize to unseen street images from another dataset.
Many qualitative examples are provided in the supplemental material.

\begin{figure*}[tb]
  \centering
  \begin{tabular}{ ccc }
  Scene & Hidden Footprints & Predicted Regions\\
  \includegraphics[width=1.5in, height=0.75in]{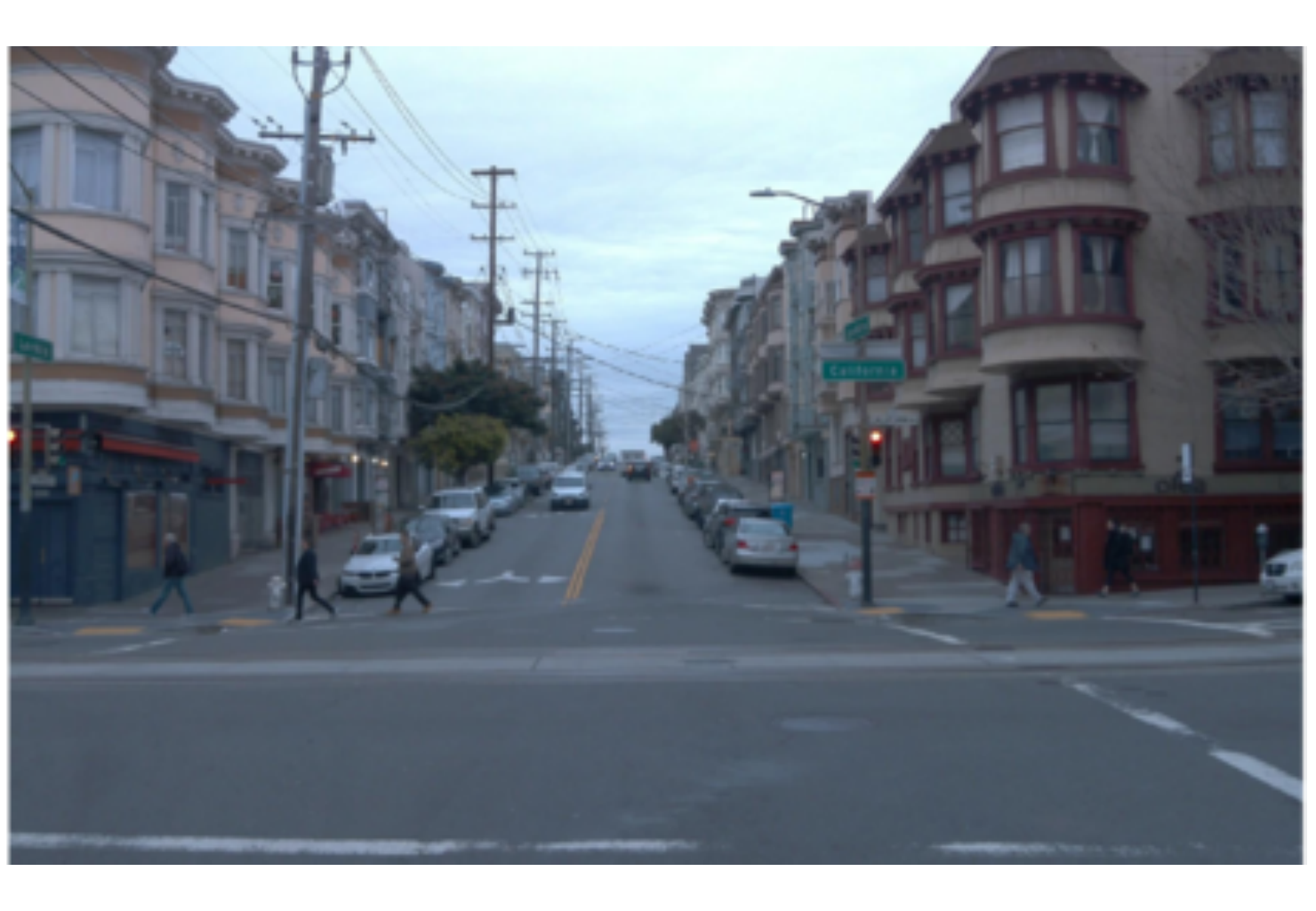} &
  \includegraphics[width=1.5in, height=0.75in]{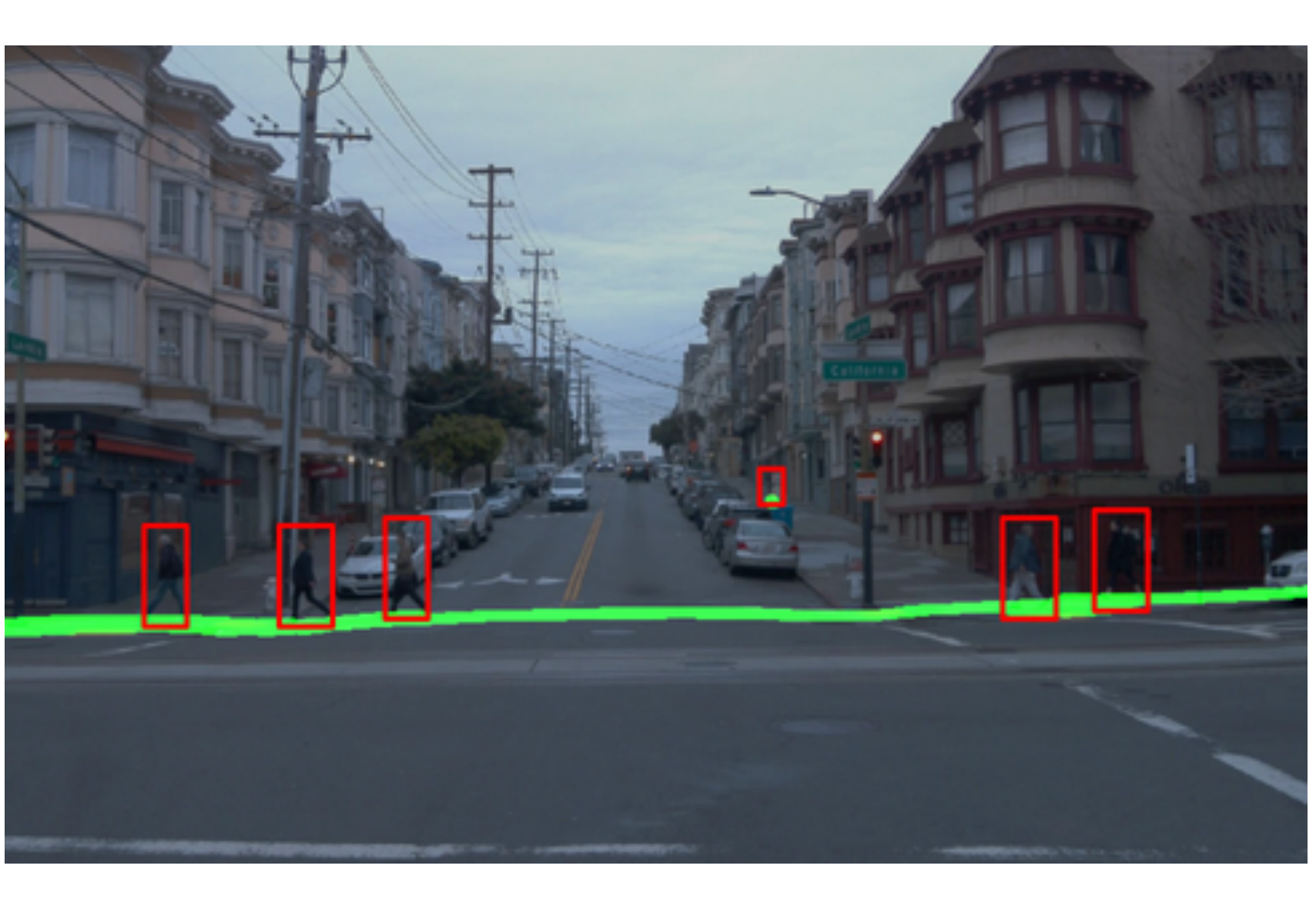} &
  \includegraphics[width=1.5in, height=0.75in]{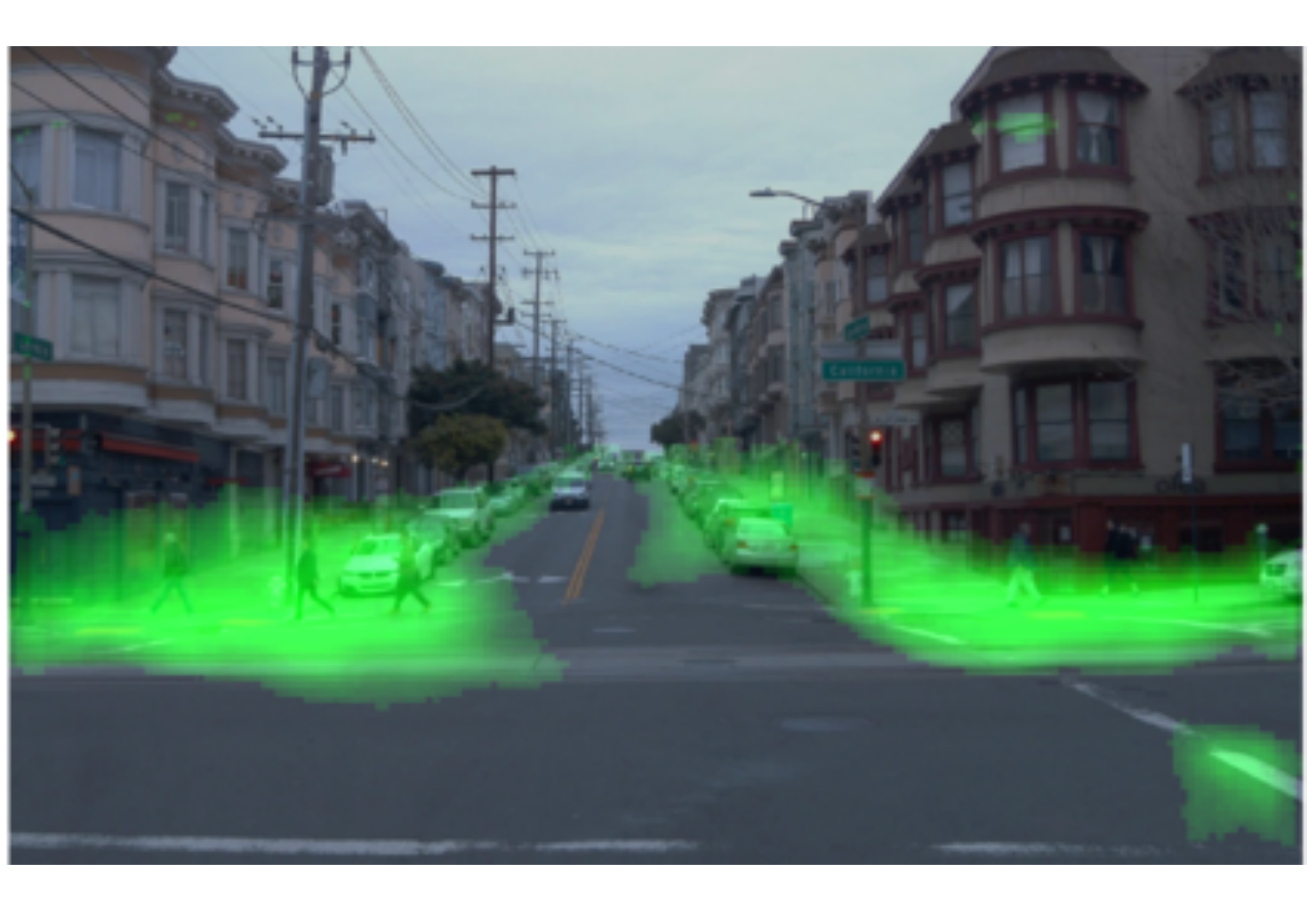}\\
   \includegraphics[width=1.5in, height=0.75in]{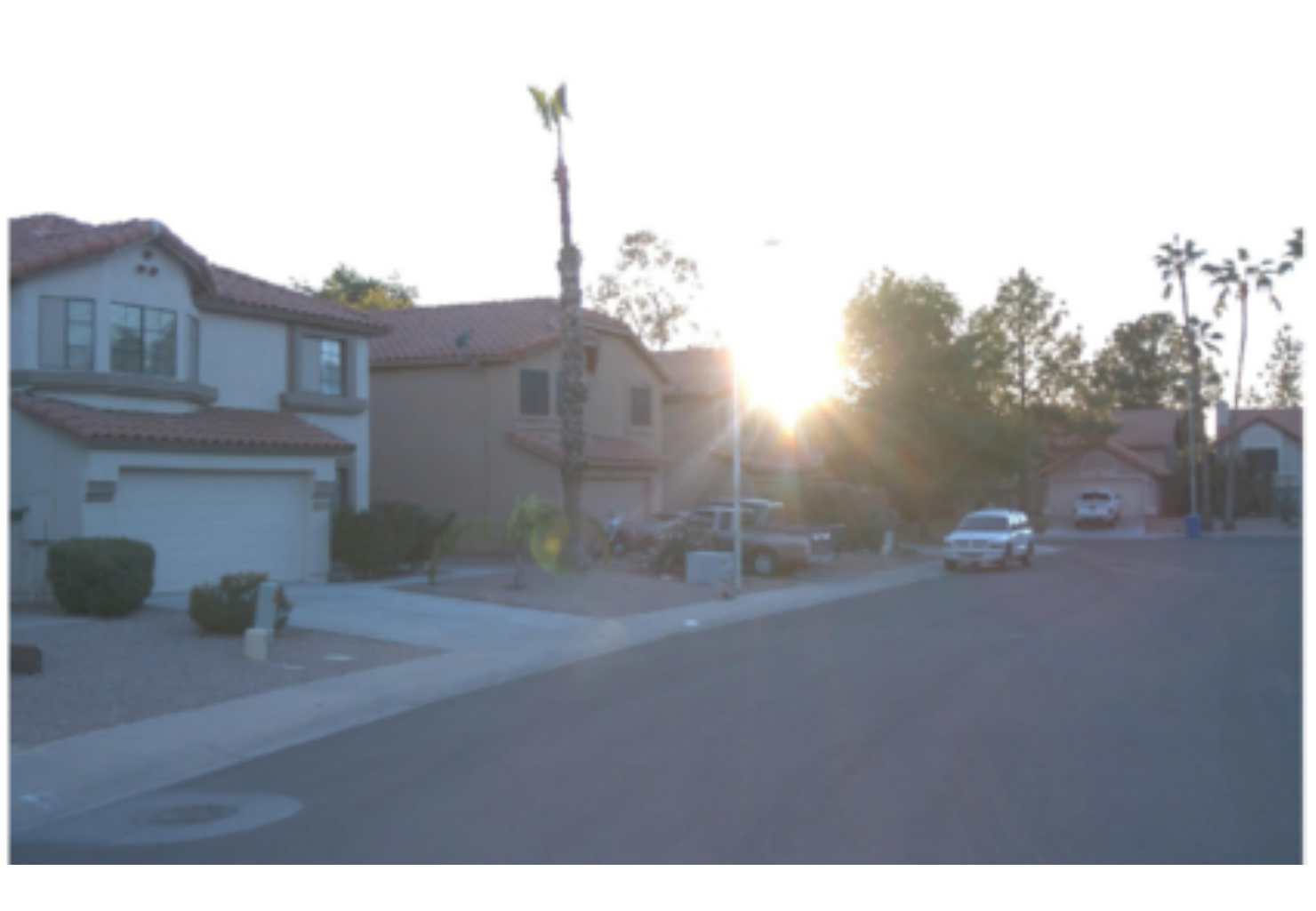} &
  \includegraphics[width=1.5in, height=0.75in]{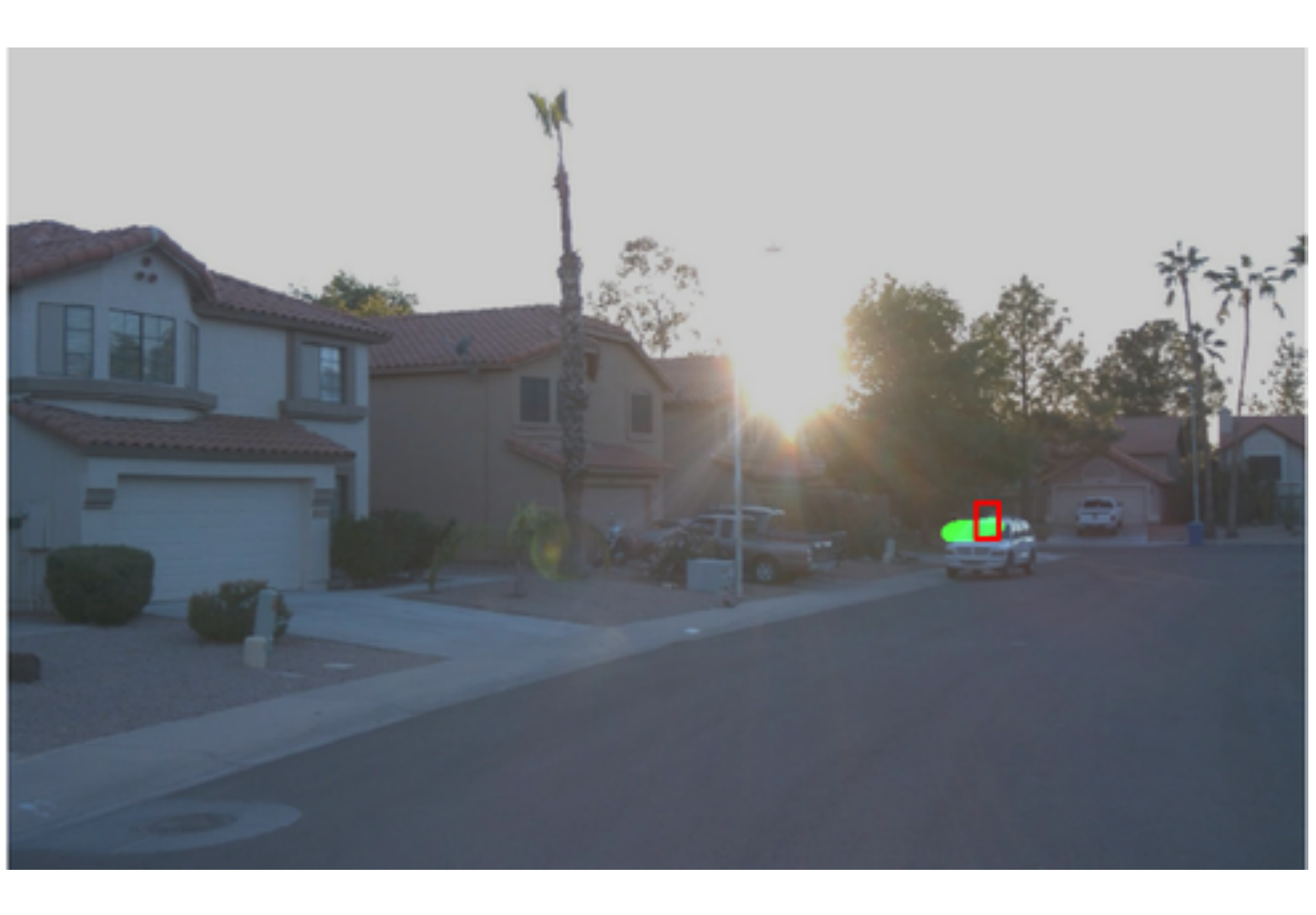} &
  \includegraphics[width=1.5in, height=0.75in]{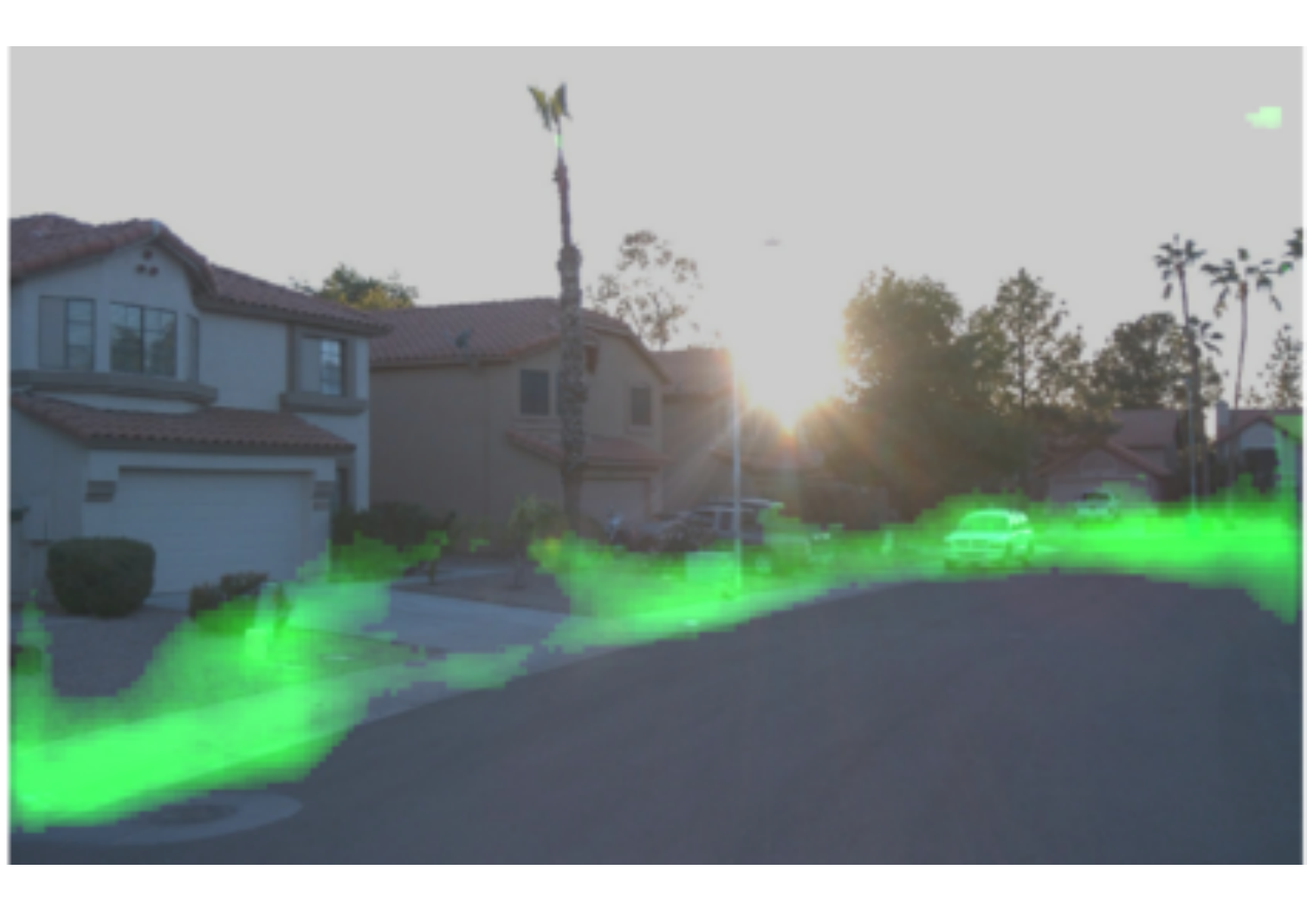}\\  \includegraphics[width=1.5in, height=0.75in]{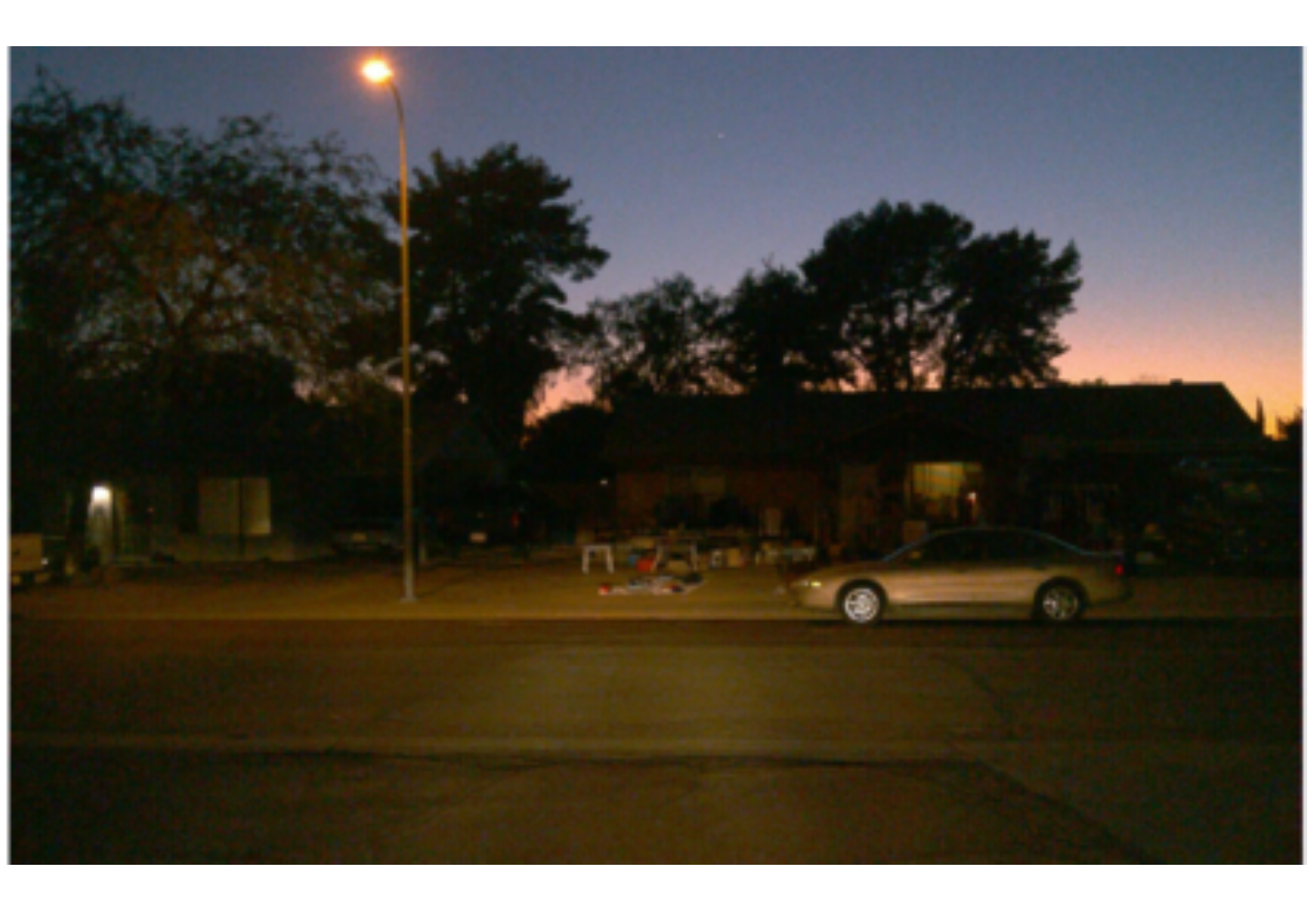} &
  \includegraphics[width=1.5in, height=0.75in]{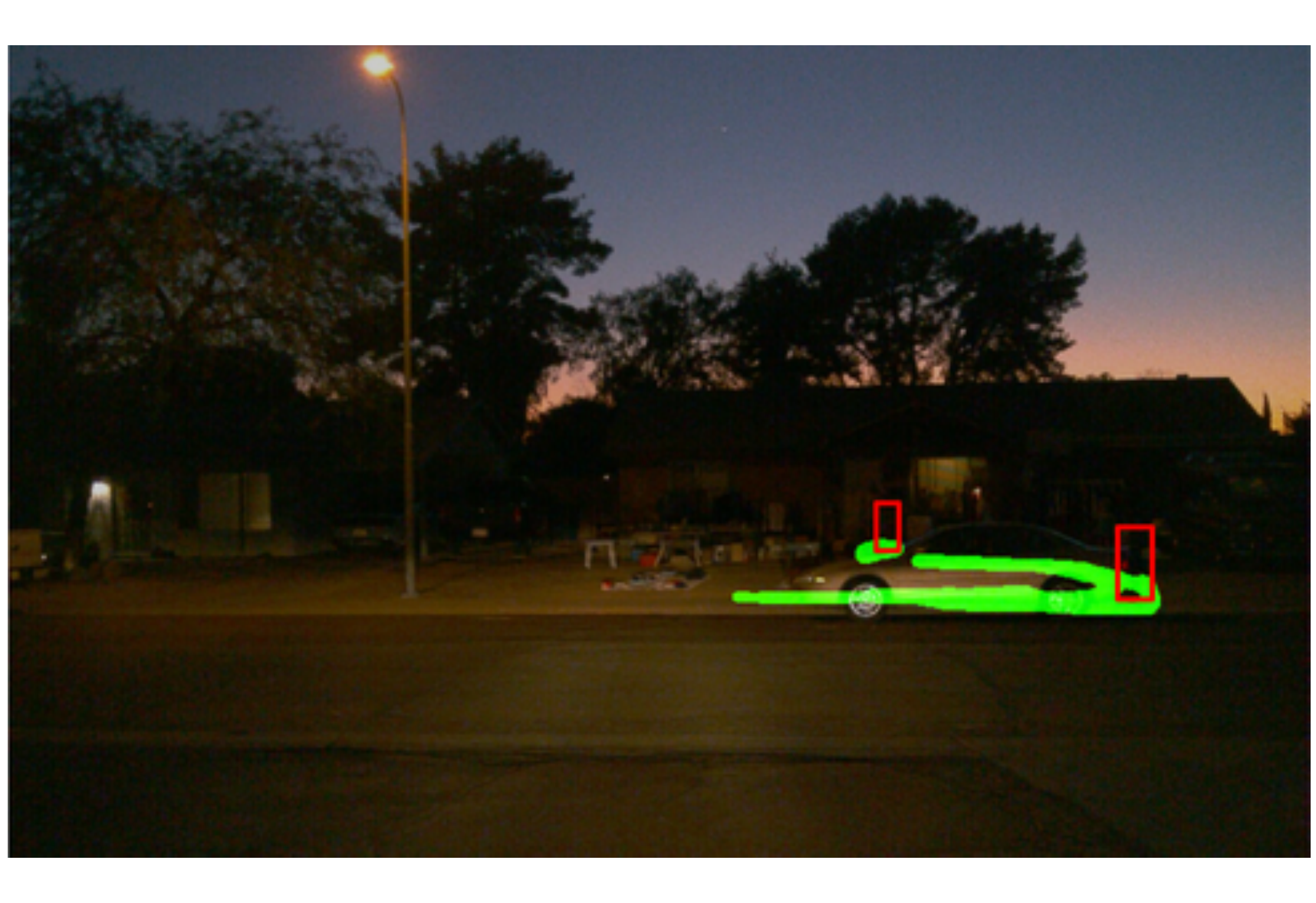} &
  \includegraphics[width=1.5in, height=0.75in]{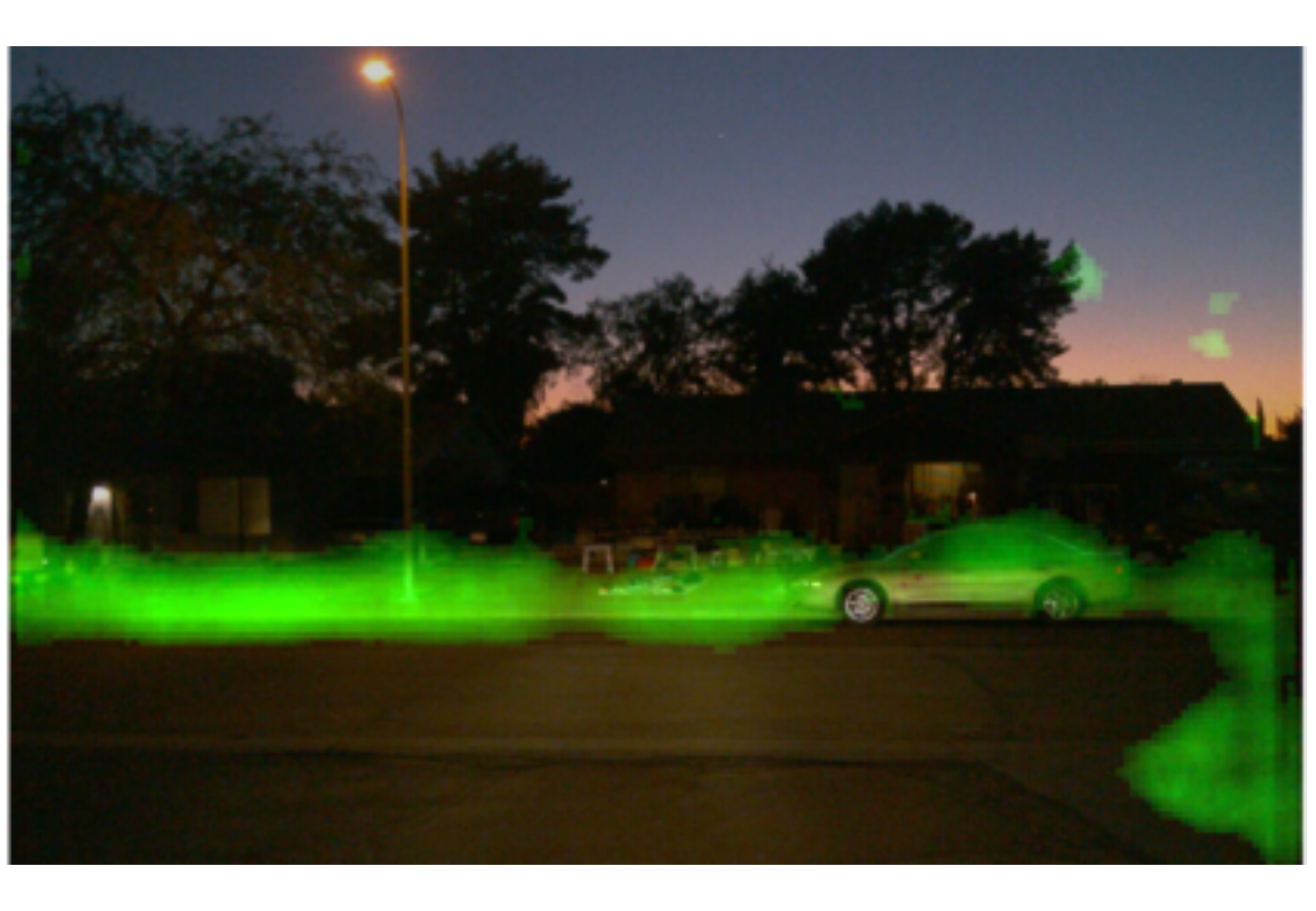}\\  \includegraphics[width=1.5in, height=0.75in]{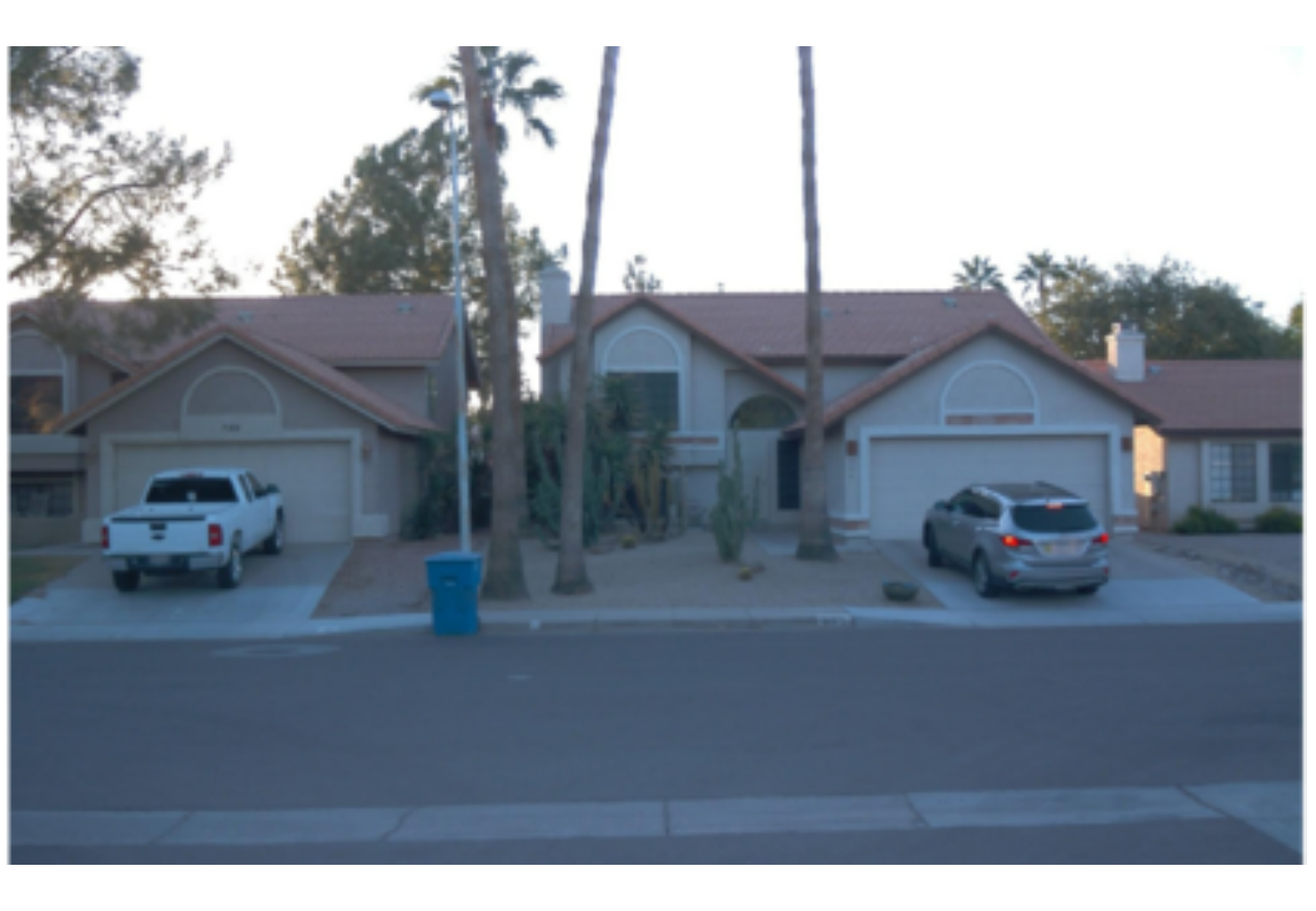} &
  \includegraphics[width=1.5in, height=0.75in]{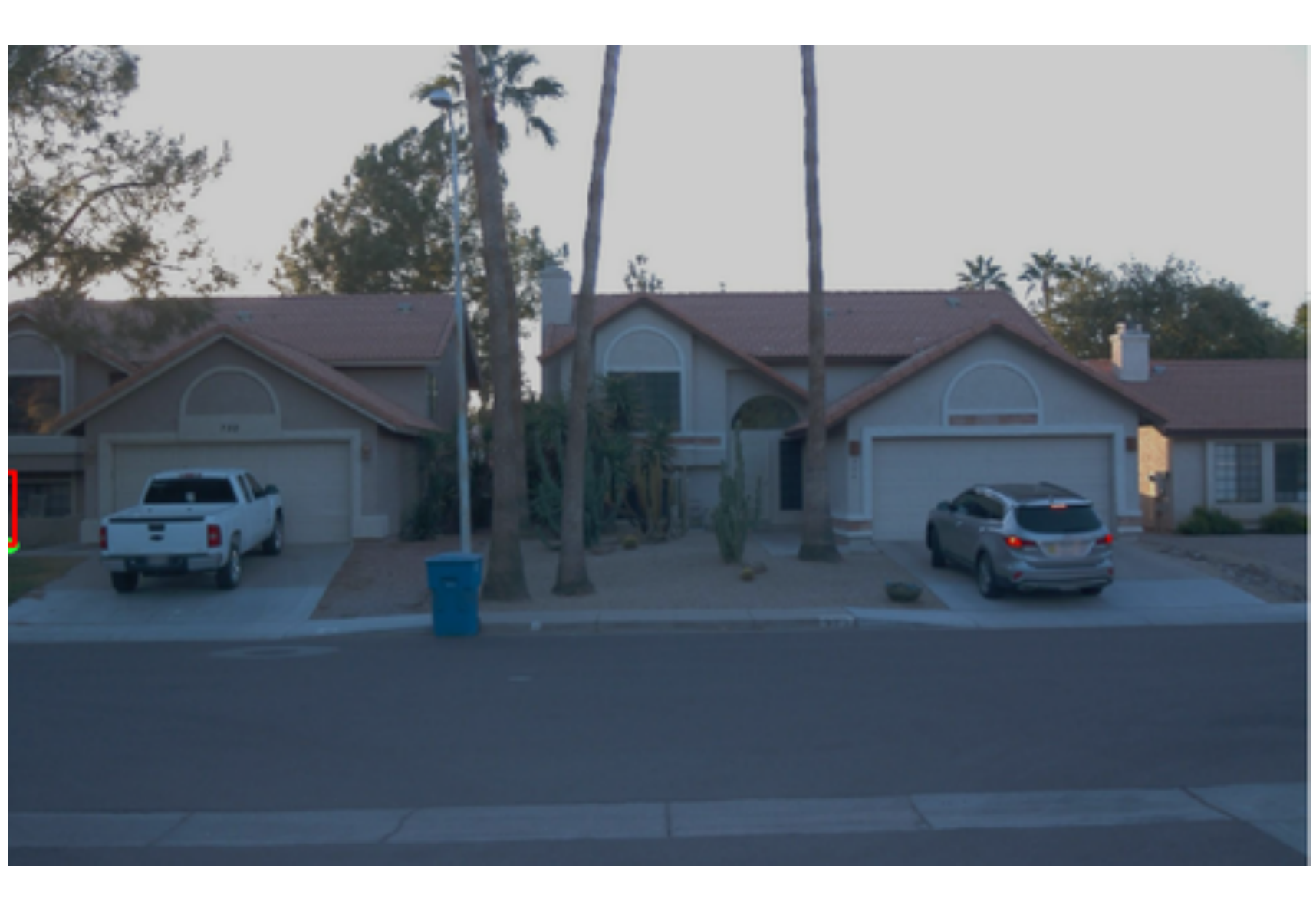} &
  \includegraphics[width=1.5in, height=0.75in]{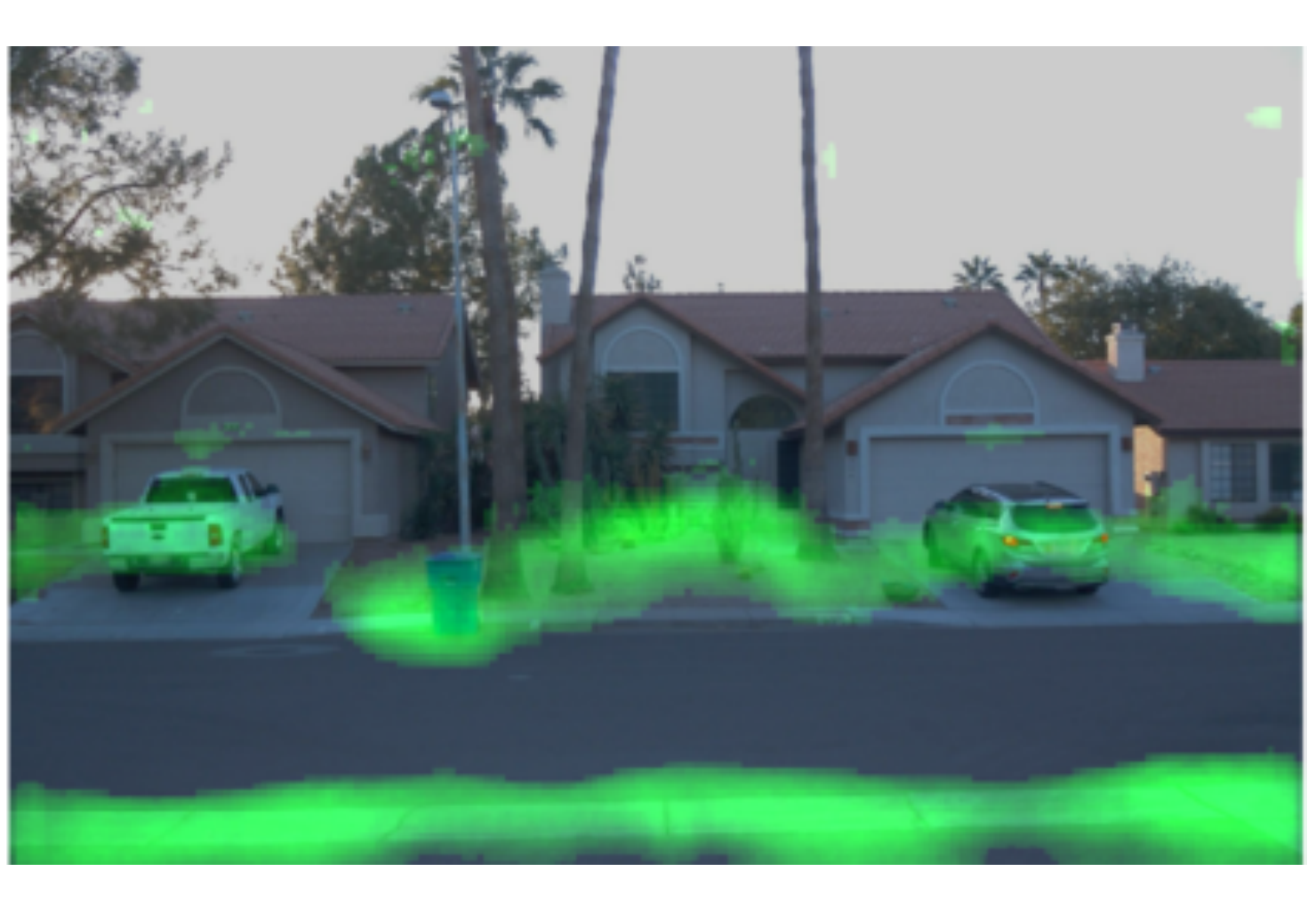}\\
\end{tabular}
  \caption{\textit{Example results on Waymo}. For each scene (left), we illustrate the projected footprints (center) and our walkability predictions (right). Red boxes are existing pedestrians. Predictions are overlaid on top of the input images for visualization purposes. Greener regions correspond to a higher likelihood for walkability. While projected footprints greatly expand the sparse existing human positions in a scene, our model predicts even more regions, yielding a nice coverage of locations such as sidewalks and crosswalks. Note that our model is not biased towards any specific configuration of predicted map, and is robust to day/night changes.}
    \label{fig:example-predicted-locs}
\end{figure*}

\subsection{Implementation details}
\noindent\textbf{Network.}
All models are trained using a generic U-Net--style dense prediction structure with an ImageNet pretrained ResNet-50 network as the backbone.
Note that our method is not limited to any particular architecture.
The network takes a $640 \times 480$ pixels image and produces a $160 \times 120$ single-channel map, where at each position a score represents the likelihood of a person being able to walk there.
Features sampled from the layer before the last single-channel output layer are used in the adversarial feature loss.
We pick the top 1\% regions and sample 10 features per image.
A three-layer MLP network is used as the discriminator, which takes a 256-D feature vector and predicts a single score.
During inference, we generate three fixed crops of an image and merge predicted maps.

\smallskip
\noindent\textbf{Training.}
We train networks using Adam with betas set to 0.5 and 0.999 and with a batch size of 10.
Images are resized to 480px in height and then randomly cropped along the width dimension to $640 \times 480$.
The initial learning rate is $10^{-4}$ and is reduced by $10\%$ when the validation loss reaches a plateau. Loss weights are $1.0$ for both the class-balanced classification loss and the adversarial feature loss.

\smallskip
\noindent\textbf{Baselines.}
We compare with models trained on hidden footprints with different losses: MSE, standard cross-entropy (BCE), Reweighted BCE, and adversarial loss--only. 
In addition, we compare with a model trained without our footprint propagation (Raw GT).
The baselines also provide insights on ablations of our design choices: our method combines the hidden footprints + class-balanced classification loss + adversarial feature loss.
All adversarial training is done with a WGAN loss with gradient penalty. 
More details are in the supplementary material.

\begin{table*}[tb]
  \begin{center}
  \setlength{\tabcolsep}{0.37em} 
  \caption{Quantitative comparisons between our proposed training (hidden footprints + class-balanced loss + adversarial feature loss) and alternatives. Best results are in bold and second best are underlined. Except Raw GT, all others are trained with propagated footprints. The numbers in the entire test set experiment are ratios, normalized by ground-truth number of locations.}
    \begin{tabular}{ lcccccc }
      \toprule
        	& Raw GT & MSE & Adversarial & BCE  & RBCE & Ours\\
        \midrule
        \multicolumn{7}{c}{Entire Test Set}\\
        \midrule
        Pred Total (TP+FP) $\uparrow$ & 0.80 & 2.12 & \textbf{22.15} & 0.24 & 3.13 & \underline{8.18}\\
        Pred Valid (TP) $\uparrow$ & 0.06 & 0.37 & \underline{0.60} & 0.13 & 0.53 & \textbf{0.69}\\
        Missing (FN) $\downarrow$ & 0.94 & 0.63 & \underline{0.39} & 0.87 & 0.47 & \textbf{0.31} \\
        Expansion  & 0.73 & 1.75 & 21.55 & 0.11 & 2.60 & 7.48\\
        \midrule
        \multicolumn{7}{c}{50-Scenes Manually Labeled Set}\\
        \midrule
        Mean Average Precision $\uparrow$ & 0.111 & 0.145 & 0.173 & 0.119 & \underline{0.190} & \textbf{0.269}\\
      \bottomrule
    \end{tabular}
  \label{tbl:waymo_expansion}
  \end{center}
\end{table*}

\subsection{Data}
The Waymo Open Dataset~\cite{waymo_open_dataset} contains over 200k images with 3D labels of four object categories, including pedestrians.
The data is divided into multiple capture sequences where each segment is a 20 second recording 
of a vehicle driving in diverse conditions.
We adopt the official train/val splits and use front camera scenes with at least one pedestrian.
Our final training and test sets contain about 80k and 2k images, respectively.
For each frame, we store 3D labels, camera intrinsics and poses.
Footprints are propagated using the procedure described in Section~\ref{sec:projection}.

\subsection{Results}

\noindent \textbf{Effect of propagating hidden footprints.}
In the raw Waymo dataset, there are on average 3.8 pedestrians per scene.
With our proposed footprints propagation, we obtain on average 362 valid locations per image, resulting in a 116$\times$ increase.
We show qualitative examples of projected labels in Figure \ref{fig:example-predicted-locs}.
While the visible pedestrians are scarce in the images, the propagated footprints have a much larger, yet still valid, coverage.

Next we show that out proposed training strategy encourages a model to make predictions that have even greater coverage over plausible regions, all from limited supervision that is severely under-sampled in the original data.

\medskip
\noindent \textbf{Quality of walkability predictions: region expansion ratio.}
Knowing that the ground-truth walkable regions are not exhaustive, a good prediction model should fully cover groundtruth positive locations, and also expand to additional locations that are also likely to be positive.
We measure the quality of such an expanded prediction by computing the following metrics on the test set:
true positives (TP), false negatives (FN), and expansion ratio compared to ground-truth.
The expansion ratio is the quotient of the number of predicted locations not covered by ground-truth and ground-truth locations.

Table \ref{tbl:waymo_expansion} top part shows that our method achieves a balance in expanding predicted regions while reducing missing regions, compared to alternative approaches.
The model trained with adversarial loss--only (Adversarial) has the largest expansion ratio as there is no other loss restricting the number of  positive predicted labels.
But the expansion ratio only tells one side of the story: how good are those expanded regions?

\begin{table*}[tb]
  \begin{center}
  \setlength{\tabcolsep}{0.37em} 
  \caption{Quantitative comparison with semantic based approaches on the 50-scenes manually labeled set. Best result is in bold.}
    \begin{tabular}{ lccc|cccc }
      \toprule
        	& Raw GT  & RBCE & Ours & S-Semantic & M-Semantic & Semantic-Net\\
        \midrule
        mAP $\uparrow$ & 0.111 & 0.190 & \textbf{0.269} & 0.087 & 0.105 & 0.201 \\
      \bottomrule
    \end{tabular}
  \label{tbl:semantic-comp}
  \end{center}
\end{table*}

\medskip
\noindent \textbf{Quality of predictions: classification metric.}
A model that predicts all regions to be positive would have the best expansion score, but that is clearly undesirable. In order to quantify whether the expanded regions are indeed good, we report standard metrics on a subset of the test data containing manually labelled positive regions.
Specifically, we labeled 50 randomly selected scenes.
For each image, we draw a walkability map of where a pedestrian might walk under normal conditions. For instance, we would not label the whole road (even though people might occasionally jaywalk).
Example labeled scenes are provided in the supplemental material.
With these fully labeled images, we can report standard classification metrics to further evaluate the quality of our predictions.

Table \ref{tbl:waymo_expansion} bottom part shows the mean Average Precision (mAP) of all methods on the manually labeled set. 
By looking at the table as a whole, it is now clear that the adversarial loss--only model produces largely false positives.
Furthermore, the benefit of the propagated footprints is clearly demonstrated by comparing MSE (0.145 mAP) to Raw GT (0.111 mAP).
Our full model significantly outperforms others (including ablated alternatives) on the manually labeled set. It shows a good balance between expansion and precision.

\begin{table*}[tb]
  \begin{center}
  \setlength{\tabcolsep}{0.37em} 
  \caption{Quantitative comparison with alternative approaches on the 100-scenes manually labeled set. Best result is in bold.}
    \begin{tabular}{ lcccc }
      \toprule
        	& Where \& Who~\cite{tan2018and}  & D-NP~\cite{chien2017detecting} & Inpainting~\cite{xie2019image} & Ours\\
        \midrule
        mAP $\uparrow$ & 0.149 & 0.142 & 0.107 & \textbf{0.246} \\
      \bottomrule
    \end{tabular}
  \label{tbl:alternative-comp}
  \end{center}
\end{table*}

\medskip
\noindent \textbf{Comparison with semantic segmentation.}
In this experiment, we test the claim that walkability predictions cannot be solved by semantic segmentation alone. 
We consider the following alternatives to our method, ordered by the richness of semantic information introduced into reasoning. 1) \textit{S-Semantic}: estimate the conditional probability of a location being walkable if it is in a single specific semantic class---we choose `sidewalk' as it is empirically the most predictive class. 2) \textit{M-Semantic}: estimate the conditional probability of a location being walkable given its predicted semantic class (from multiple classes). 3) \textit{Semantic-Net}: given a semantic segmentation as input, train a neural network to predict walkability. 

Table \ref{tbl:semantic-comp} shows mAP scores on the manually labeled set. 
We use a state-of-the-art semantic segmentation network model, HRNet~\cite{HRNet1_SunXLW19,HRNet2_WangSCJDZLMTWLX19}, that is pre-trained on street scenes. 
The generated semantic maps are reasonable upon visual inspection.
All semantics-based approaches are trained with our propagated footprints.
The clear performance advantage of our method suggests that predicting walkability is not trivially a downstream application of semantic segmentation. 
An interesting future direction is to explore combining our approach with semantics.

\medskip
\noindent \textbf{Comparison with alternative walkability predictions.} Here, we compare to three alternative walkability prediction methods:

\smallskip
\noindent\textit{Where \& Who \cite{tan2018and}:} This method generates person compositions given an RGB image and its semantic layout. We use the authors' provided model trained on the COCO dataset \cite{coco}.
The semantic layouts are obtained from Detectron2 \cite{wu2019detectron2}.

\smallskip
\noindent\textit{D-NP \cite{chien2017detecting}:} This method predicts human poses from empty scenes. We adapt it to predict a heatmap of people's feet. We use the same U-Net structure as ours for fair comparison. D-NP has an additional discriminator to classify (scene, map) pairs, adopted from DCGAN as in \cite{chien2017detecting}. We train the whole pipeline on the Waymo dataset with inpainted scenes in which people are removed.

\smallskip
\noindent\textit{Inpainting \cite{xie2019image}:} For a stronger baseline, we use a state-of-the-art image inpainting algorithm pre-trained on the Paris StreetView data \cite{paris} to remove existing people. The original locations of people are then used as the ground-truth walkable regions. With the empty scenes and walkable regions, we train a regression network with the same U-Net structure as ours and with a 2D binary cross-entropy loss.

\medskip
\noindent Table \ref{tbl:alternative-comp} shows results on our Waymo manually labeled set, for which we include an additional set of scenes such that there are in total 100 scenes with walkability maps. 
Our full model outperforms all others by a significant margin.

\begin{figure*}[tb]
  \centering
  \begin{tabular}{ ccc }
  Input image & Prediction  of Lee \etal~\cite{lee2018context} & Our Prediction\\
  \includegraphics[width=1.5in, height=0.75in]{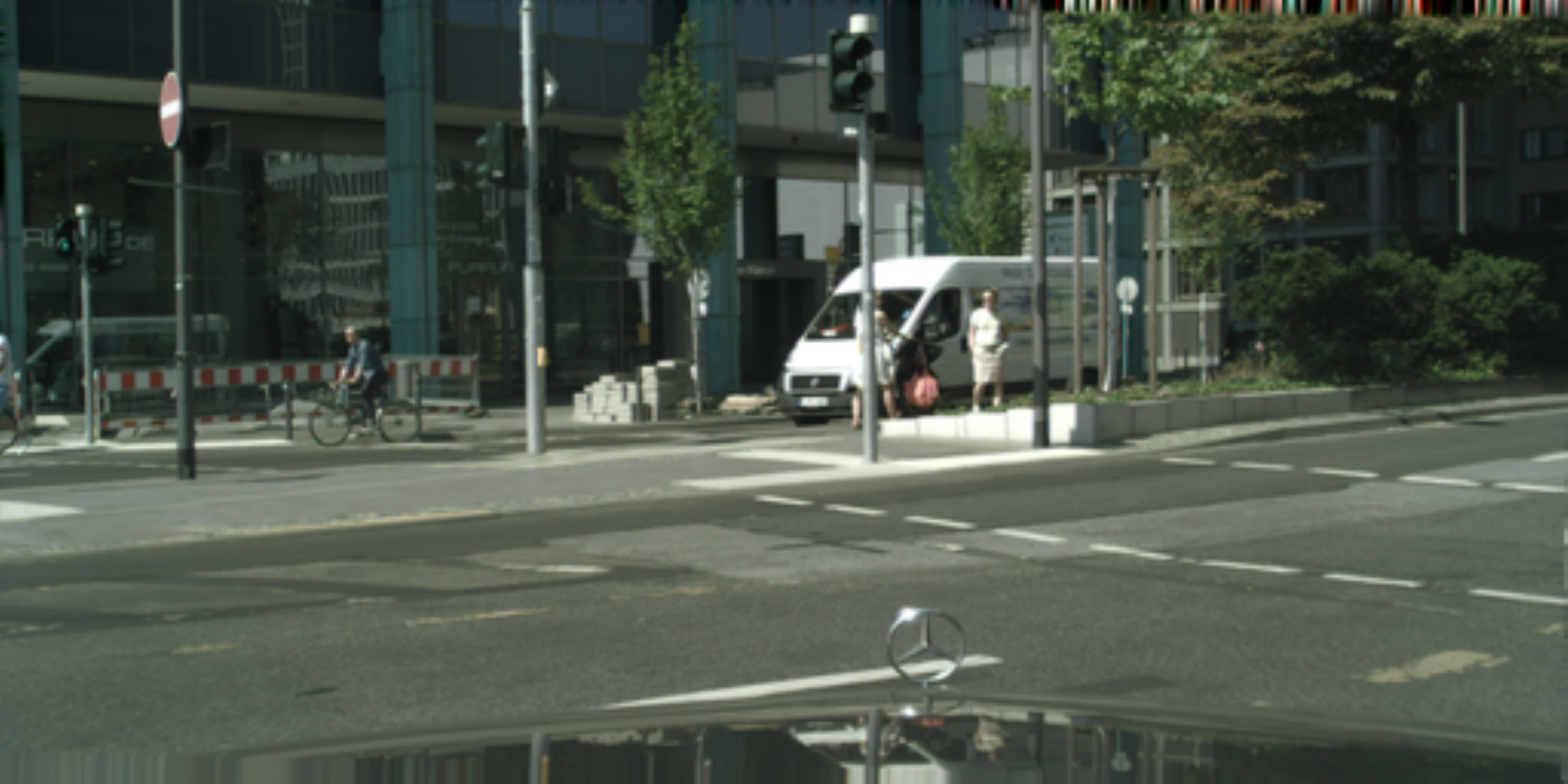} &
  \includegraphics[width=1.5in, height=0.75in]{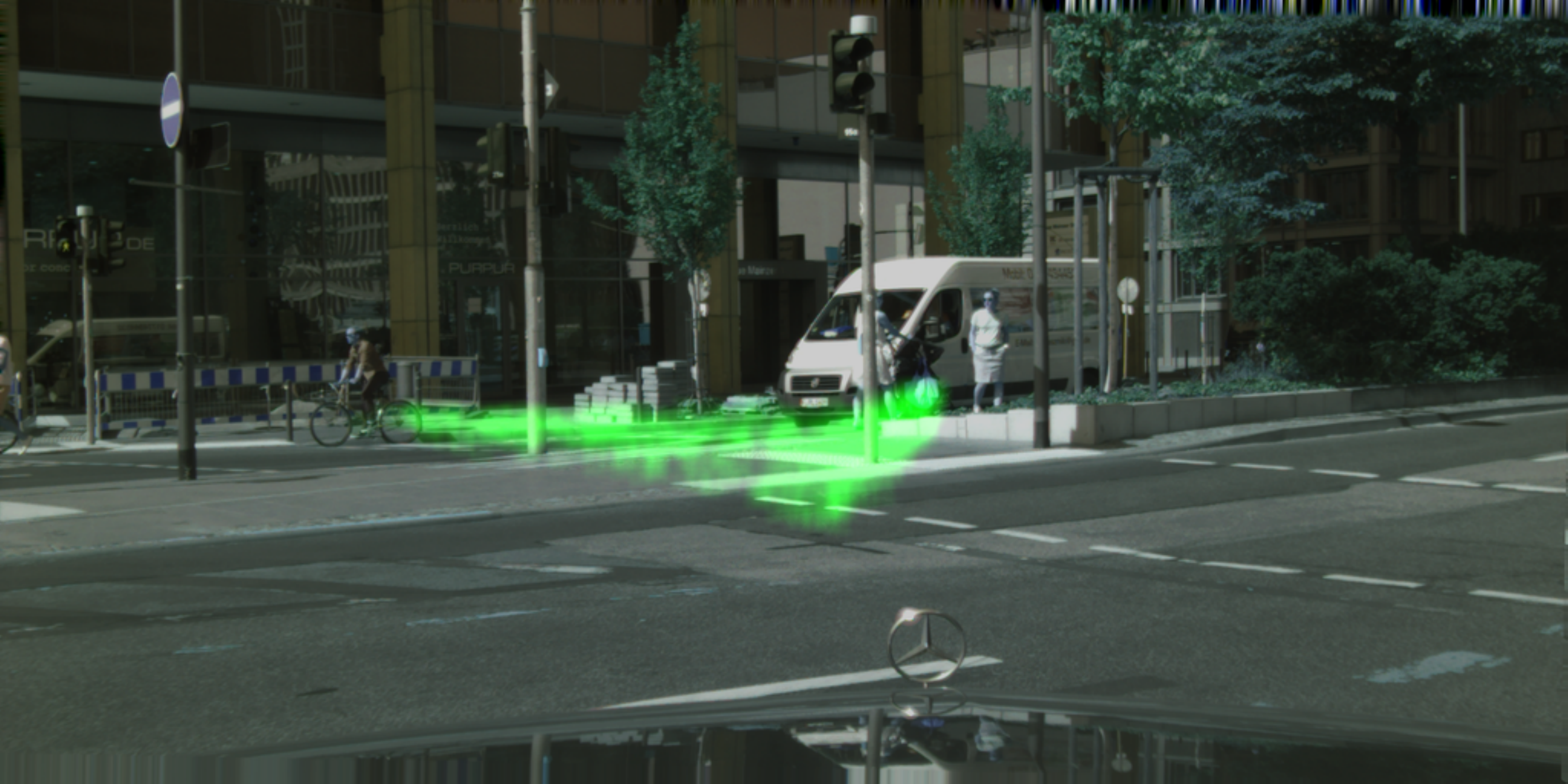} &
  \includegraphics[width=1.5in, height=0.75in]{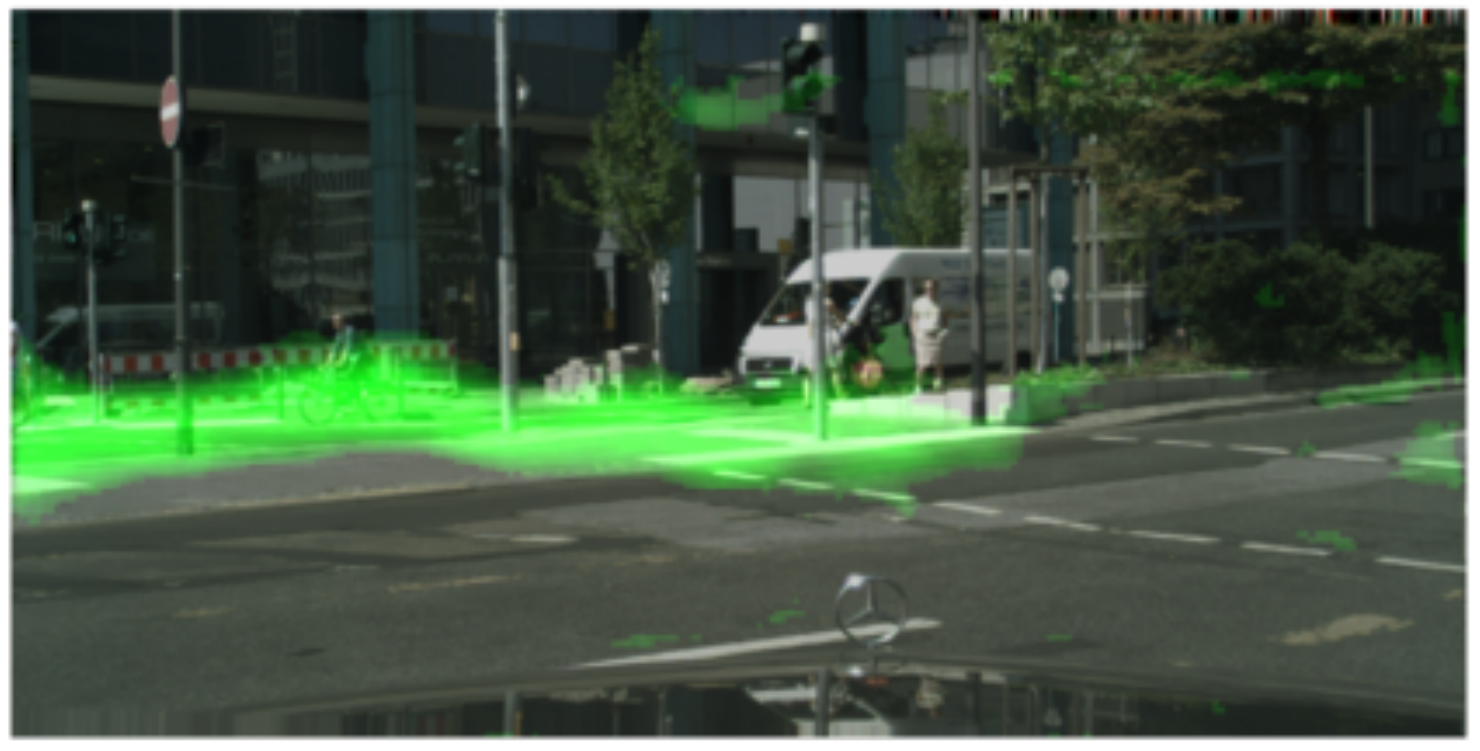}\\
  \includegraphics[width=1.5in, height=0.75in]{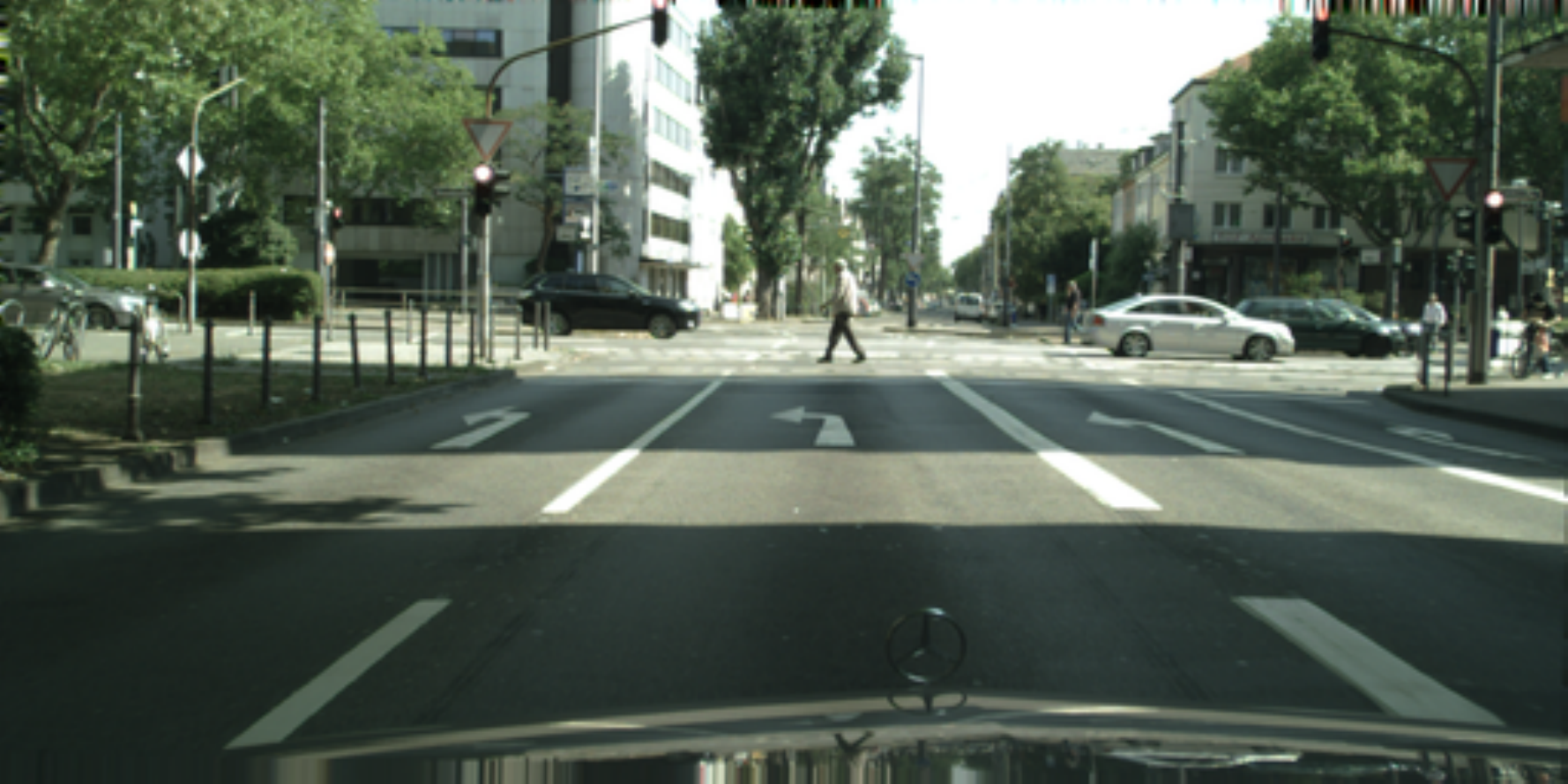} &
  \includegraphics[width=1.5in, height=0.75in]{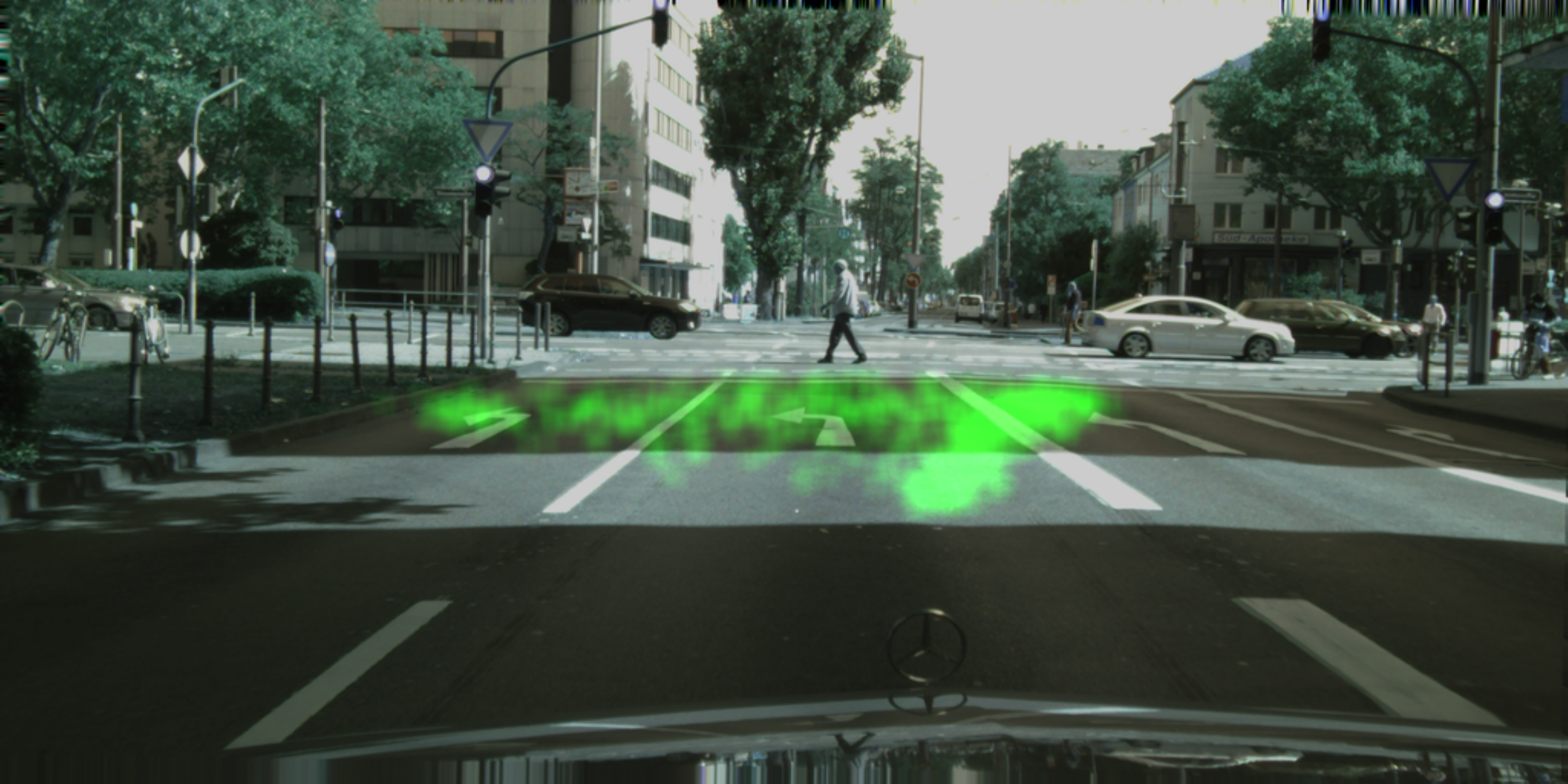} &
  \includegraphics[width=1.5in, height=0.75in]{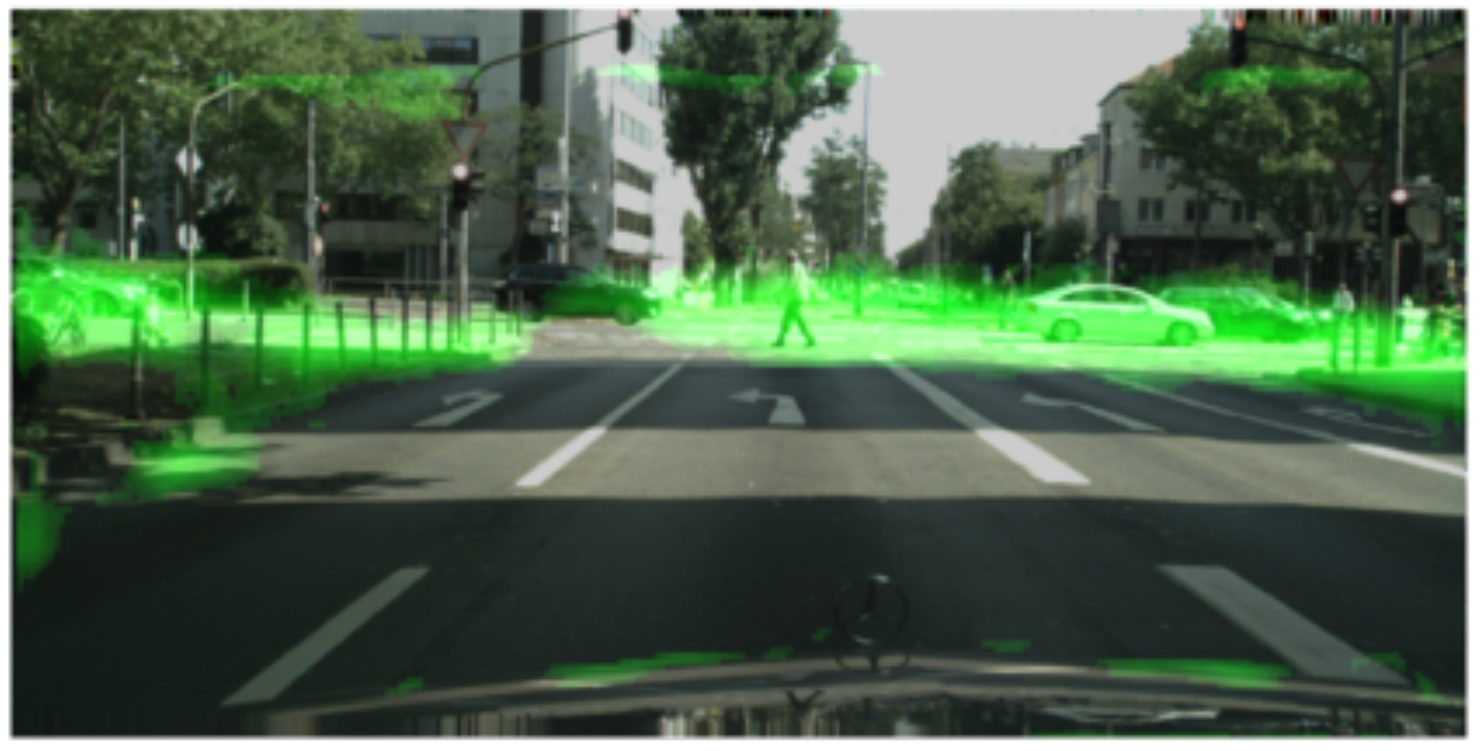}\\
  \includegraphics[width=1.5in, height=0.75in]{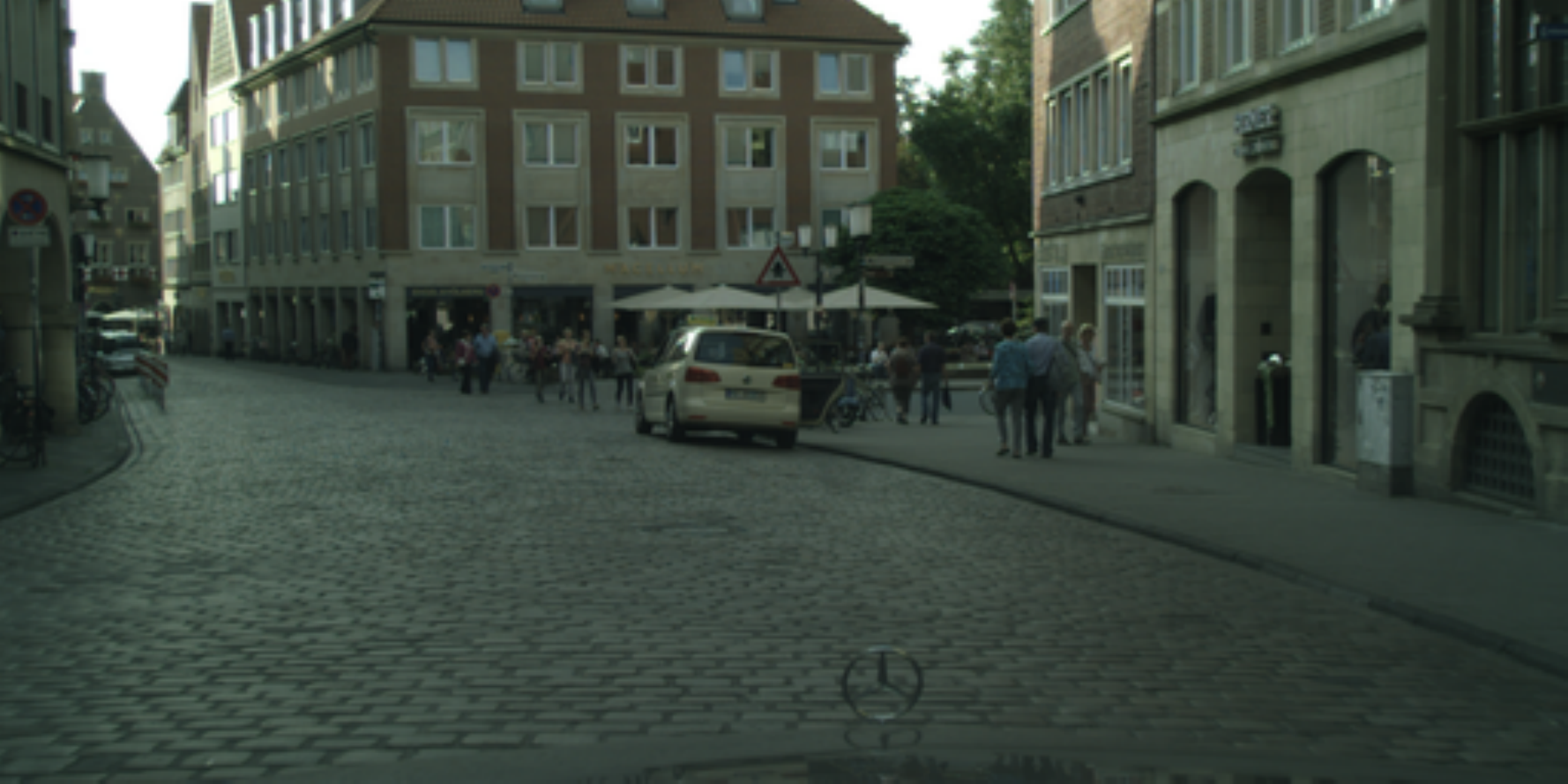} &
  \includegraphics[width=1.5in, height=0.75in]{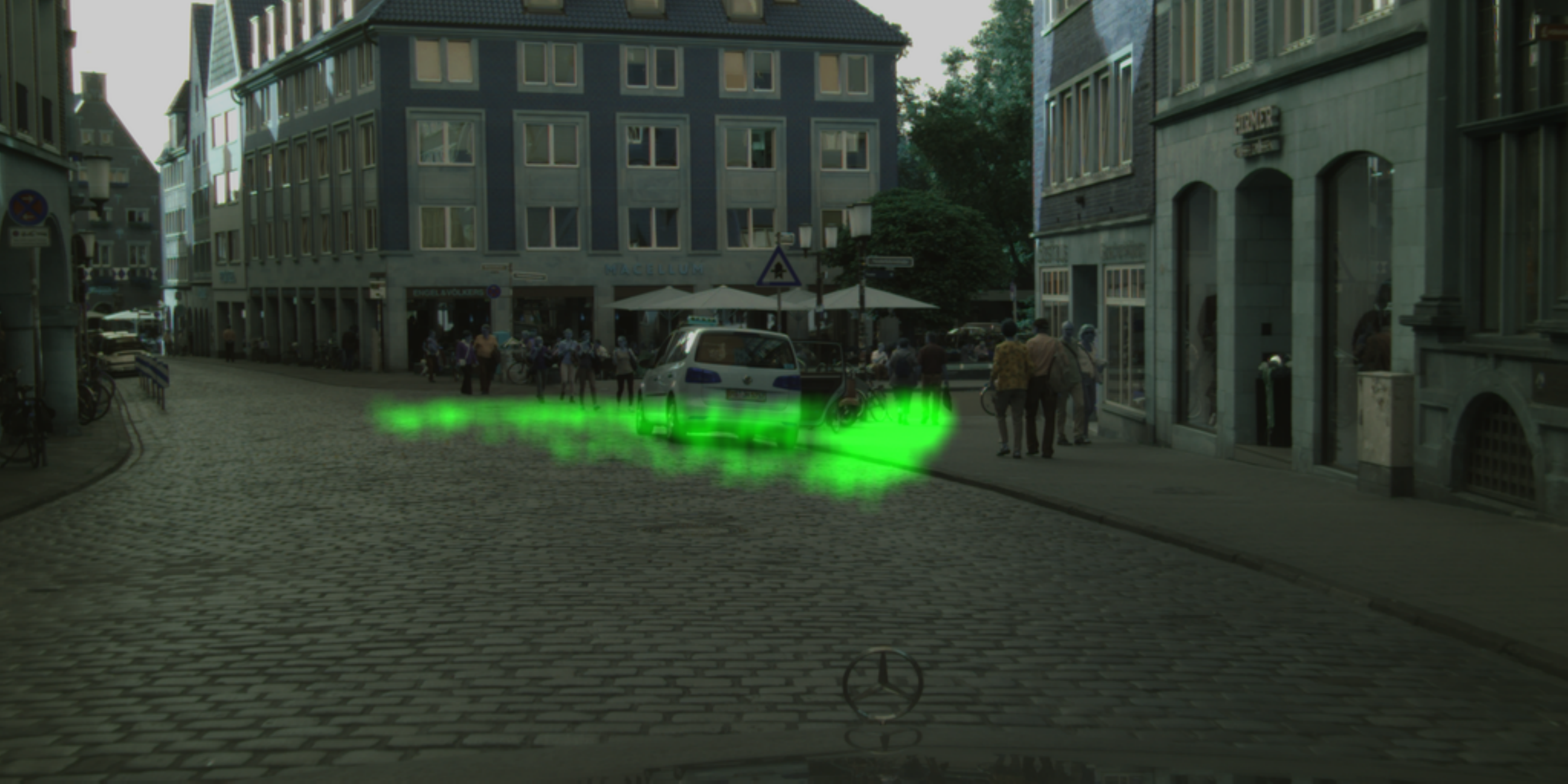} &
  \includegraphics[width=1.5in, height=0.75in]{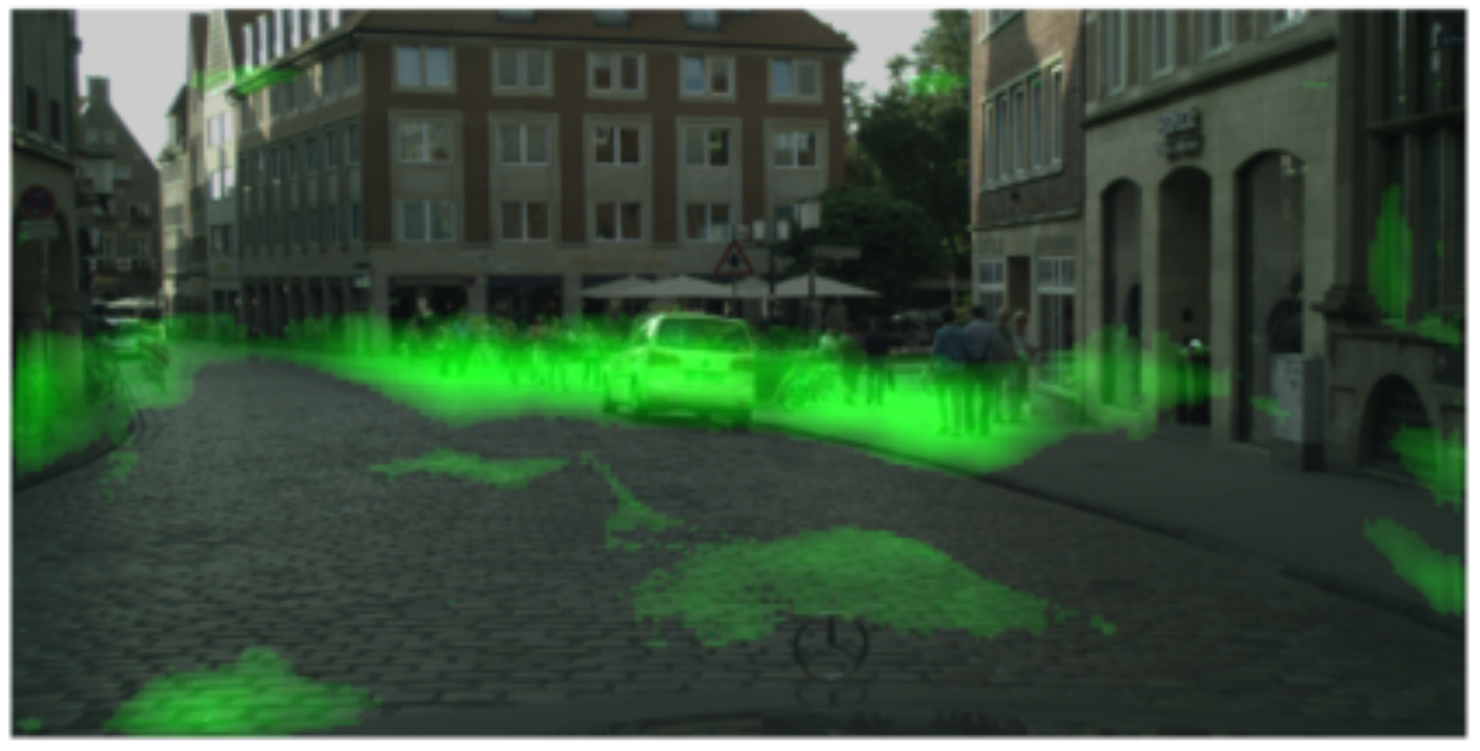}\\
\end{tabular}
  \caption{\emph{Generalization to Cityscapes}. This figure compares predicted walkability maps produced by a semantics-based pedestrian insertion method \cite{lee2018context} and by our Waymo-trained model on the Cityscapes dataset.
  Locations predicted by our method are diverse and extensive.
  }
    \label{fig:example-cityscapes}
\end{figure*}

\medskip
\noindent \textbf{Generalization to Cityscapes.}
We also evaluate our Waymo-trained model on the Cityscapes datasets \cite{cordts2016cityscapes} \textit{without finetuning}, demonstrating its generalization ability. 
Cityscapes contains $5,000$ images and
we evaluate on the official validation set.
In addition to RGB images, the dataset also provides a fine-grained semantic label map of common classes (e.g., building, bicycle, and ground).

We use a semantics-based metric to measure the similarity between our predicted walkable locations and the ground-truth person locations. 
To this end, we sample pedestrian feet locations both from our predicted map, as well as from the ground-truth. 
At each location, we accumulate the local surrounding areas' ground-truth semantic labels and store the most frequent label as where the person is standing on.
A histogram of those labels is collected by sampling locations over all images in the validation set. This shows distribution of 
walkable regions according to semantic categories.
For comparison, we also sample predicted person locations predicted from Lee \etal~\cite{lee2018context}, which was trained on the Cityspaces dataset, and collect a histogram from those locations. 
Note that unlike in \cite{lee2018context}, our method does not require semantic maps as inputs.

To measure the discrepancy of the distributions between the predicted walking locations and ground-truth, we compute the Kullback–Leibler (KL) divergence. 
The KL-divergence between our model (trained on Waymo, and not fine-tuned on Cityscapes) and the ground-truth is $0.42$. This distance is  significantly smaller than the distance of $0.78$ achieved by the Cityscapes-trained model of \cite{lee2018context}.
Figure \ref{fig:example-cityscapes} shows qualitative comparisons.
Taking a closer look at a few selected semantic categories, 
$50\%$ of the predictions from the model proposed in \cite{lee2018context} are in the road,
whereas our method places $18\%$ of the people in streets, compared to $21\%$ of the ground-truth people.
In contrast, our model predicts $26\%$ of the pedestrians on regions already covered by cars, 
although in ground-truth  $11\%$ of people has feet locations on car segments
(and $8\%$ in \cite{lee2018context}). We attribute this behaviour to the fact that people can be walking behind cars in our training data, thus our model may predict occluded regions. 
In the supplemental material we show a full distribution of the semantic categories associated with our walkability maps vs.\ those of ground-truth labeled people.

\subsection{Predicting walking directions}
\begin{figure*}[tb]
\centering
\includegraphics[width=\textwidth]{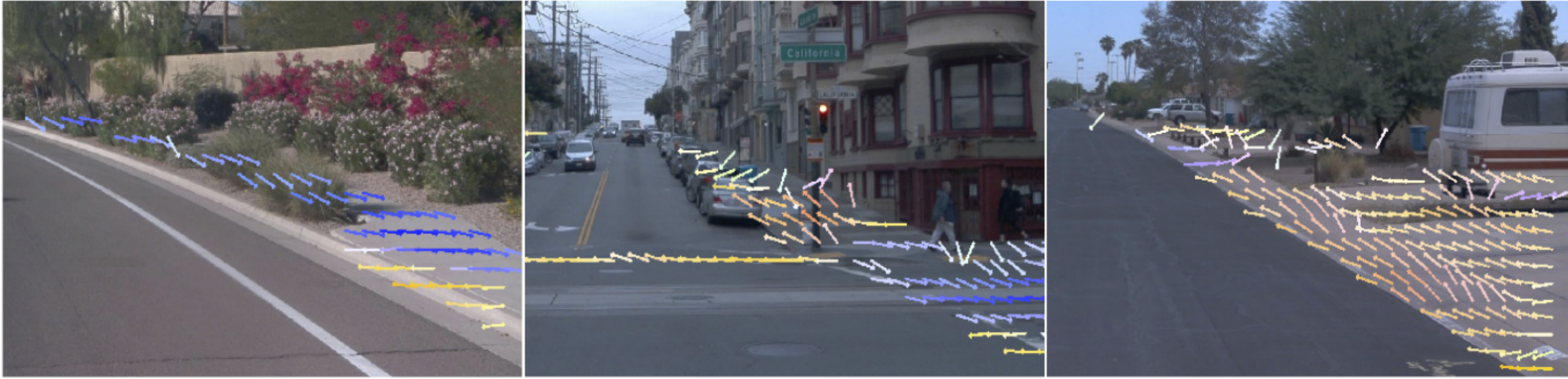}\\
\caption{Predicted walking directions in Waymo scenes on valid walkable regions. Directions are colored following a common optical flow visualization scheme~\cite{opticalflow}.}
\label{fig:flows}
\end{figure*}

So far we have focused on modeling the \emph{locations} where people walk in a scene.
Here, we show that our hidden footprint propagation setting is adaptable to a richer family of predictions.
In particular, we show a preliminary study of predicting a person's walking \emph{directions}.
For each person in the dataset, we track their trajectory, compute 3D walking directions at each time step, and then project this direction to all frames using the procedure described in Section~\ref{sec:projection}. 
We train a neural network to predict a pixel-wise map where at each location it outputs a unit-length directional vector.
Figure~\ref{fig:flows} shows qualitative results of our predicted walking directions.

\section{Conclusion}

We proposed a 
new method for predicting where people can potentially walk in a scene.
To circumvent the challenges of insufficient labels, we first project existing human labels from each frame to each other frame by utilizing 3D information.
We then devise a training strategy to encourage the network to learn the true underlying distribution from under-sampled ground-truth labels.
Our experiments show that our model can expand ground-truth human walkable locations in the Waymo Open Dataset, and can also outperform state-of-the-art human context models, including on unobserved datasets such as Cityscapes.

\medskip
\noindent \textbf{Acknowledgements.} This research was supported in part by the generosity of Eric and Wendy Schmidt by recommendation of the Schmidt Futures program.

\bibliographystyle{splncs04}
\bibliography{mybib}

\end{document}